\documentclass[runningheads]{llncs}

\usepackage{eccv}

\usepackage{eccvabbrv}

\usepackage{graphicx}
\usepackage{booktabs}

\usepackage[accsupp]{axessibility}  % Improves PDF readability for those with disabilities.

\usepackage{algorithm}
\usepackage{algpseudocode}
\usepackage{bbm}
\usepackage{ulem}

\usepackage{multirow}
\usepackage{booktabs,arydshln}
\makeatletter
\def\adl@drawiv#1#2#3{%
        \hskip.5\tabcolsep
        \xleaders#3{#2.5\@tempdimb #1{1}#2.5\@tempdimb}%
                #2\z@ plus1fil minus1fil\relax
        \hskip.5\tabcolsep}
\newcommand{\cdashlinelr}[1]{%
  \noalign{\vskip\aboverulesep
           \global\let\@dashdrawstore\adl@draw
           \global\let\adl@draw\adl@drawiv}
  \cdashline{#1}
  \noalign{\global\let\adl@draw\@dashdrawstore
           \vskip\belowrulesep}}
\makeatother

\usepackage{pifont}
\newcommand{\cmark}{\ding{51}}%
\newcommand{\xmark}{\ding{55}}%

\def\with{\textbf{w/}\xspace}
\def\without{\textbf{w/o}\xspace}

\newcommand{\algname}{\textit{Scissorhands}\xspace}
\usepackage{amsmath,amsfonts,bm, mathtools}

\def\eqref#1{equation~\ref{#1}}
\def\1{\bm{1}}

\def\rvm{{\mathbf{m}}}

\def\rvx{{\mathbf{x}}}
\def\rvy{{\mathbf{y}}}

\def\vzero{{\bm{0}}}
\def\vone{{\bm{1}}}

\def\vtheta{{\bm{\theta}}}
\def\va{{\bm{a}}}

\def\ve{{\bm{e}}}

\def\vg{{\bm{g}}}

\def\vx{{\bm{x}}}
\def\vy{{\bm{y}}}

\def\mA{{\bm{A}}}

\def\mX{{\bm{X}}}

\DeclareMathAlphabet{\mathsfit}{\encodingdefault}{\sfdefault}{m}{sl}
\SetMathAlphabet{\mathsfit}{bold}{\encodingdefault}{\sfdefault}{bx}{n}

\def\gD{{\mathcal{D}}}

\def\gN{{\mathcal{N}}}

\def\gU{{\mathcal{U}}}

\def\sR{{\mathbb{R}}}

\newcommand{\R}{\mathbb{R}}

\DeclareMathOperator*{\argmin}{arg\,min}

\usepackage{hyperref}

\usepackage{orcidlink}

\begin{document}

\title{Scissorhands: Scrub Data Influence via Connection Sensitivity in Networks} 

\titlerunning{Scissorhands: Scrub Data Influence via Connection Sensitivity in Networks}

\author{Jing Wu\textsuperscript{\rm 1}\orcidlink{0009-0004-7049-5480}, Mehrtash Harandi\textsuperscript{\rm 1}\orcidlink{0000-0002-6937-6300}}

\authorrunning{Jing Wu, Mehrtash Harandi}

\institute{
\textsuperscript{\rm 1}Department of Electrical and Computer Systems Engineering \\ 
Monash University, Australia \\
\email{\{jing.wu1, mehrtash.harandi\}@monash.edu}
}

\maketitle

\begin{abstract}
Machine unlearning has become a pivotal task to erase the influence of data from a trained model. It adheres to recent data regulation standards and enhances the privacy and security of machine learning applications.
In this work, we present a new machine unlearning approach \algname.
Initially, \algname identifies the most pertinent parameters in the given model relative to the forgetting data via connection sensitivity. By reinitializing the most influential top-$k$ percent of these parameters, a trimmed model for erasing the influence of the forgetting data is obtained.
Subsequently, \algname fine-tunes the trimmed model with a gradient projection-based approach, seeking parameters that preserve information on the remaining data while discarding information related to the forgetting data.
Our experimental results, conducted across image classification and image generation tasks, demonstrate that \algname, showcases competitive performance when compared to existing methods. Source code is available at \url{https://github.com/JingWu321/Scissorhands}.


\subsubsection{Warning:} This paper contains explicit sexual imagery that may be offensive.

\end{abstract}
\section{Introduction}
\label{sec:intro}

In this work, we aim to propose an effective machine unlearning method.
Under data regulations like the European Union General Data Protection Regulation (GDPR)~\cite{voigt2017eu} and the California Consumer Privacy Act (CCPA)~\cite{goldman2020introduction}, all users are granted the \textit{right to be forgotten}.
In machine learning, these legal provisions empower data owners with the right not only to withdraw their data from trained models but also to ensure that their data's influence on these models is erased.

The most direct approach to accomplish this objective is to retrain the model from scratch, excluding the data requested for deletion from the training process. Retraining from scratch is typically considered the gold standard in the field of machine unlearning~\cite{thudi2022unrolling,fan2023salun}.
Yet, this poses a challenge as numerous in-production models require prolonged training periods and substantial computing resources. While retraining is feasible, it is often impractical.
Consequently, the proposal and development of efficient approximate unlearning methods~\cite{guo2019certified,tarun2023fast, tarun2023deep, foster2023fast, heng2023selective} have become essential.
Currently, most approximate unlearning techniques achieve forgetting by adding normally distributed noise to the parameters~\cite{golatkar2020eternal,golatkar2021mixed,golatkar2020forgetting} or by estimating the influence of a particular data point
on the model’s predictions~\cite{guo2019certified,neel2021descent,sekhari2021remember,mehta2022deep} based on the influence function method~\cite{dennis1982residuals}.
Jia \etal~\cite{jia2023model} demonstrate that model sparsity can help to unlearn and fuse model sparsity into the unlearning process.

Fan \etal~\cite{fan2023salun} recently raised concerns regarding the instability in approximate unlearning methods, and highlighted that the current machine unlearning approaches, initially crafted for image classification tasks, fall short in effectively tackling the challenges of machine unlearning within the realm of image generation.
The authors introduce a novel unlearning method SalUn that can effectively perform forgetting in both image classification and generation tasks. Their unlearning mechanism includes finding salient weights and then fine-tuning these weights using forgetting data assigned with random labels.
However, studies like \cite{zhang2021understanding} show that the model can still learn the true data distribution even using data with random labels.
While SalUn has shown state-of-the-art performance, it could potentially memorize information about forgetting data points, albeit associating them with random labels but still can be undesired in sensitive applications.
In \textsection~\ref{sec:experiment}, results also show that SalUn tends to memorize knowledge about the forgetting data.

\noindent
\textbf{This work.}
We present an unlearning approach, \algname, designed to erase data influence in the classification models and eliminate a particular concept from a text-to-image model.
To achieve this, inspired by~\cite{jia2023model}, our key insight is to first erase the critical influence of forgetting data in models and then relearn.
Unlike previous methods that employ the forgetting data with noisy labels as a part of relearning, we propose to use an efficient gradient projection method to relearn the critical features and patterns while ensuring the exclusion of influences associated with the forgetting data.
Through a series of experiments and evaluations, including the classification on SVHN~\cite{netzer2011reading}, CIFAR-10 and CIFAR-100~\cite{krizhevsky2009learning}, CelebAMASK-HQ~\cite{lee2020maskgan} datasets, as well as the open-source stable diffusion~\cite{rombach2022high} text-to-image model, results demonstrate the viability and effectiveness of our technique in forgetting the influence of random samples, discrete classes, and sensitive content such as nudity.

\section{Methodology}
\label{sec:method}

In this section, we propose \algname, our unlearning framework that scrubs data from a model by eliminating their influence on model parameters.
Throughout the paper, we denote scalars and vectors/matrices by lowercase and bold symbols, respectively (\eg, $a$, $\va$, and $\mA$).
We show a $d$-dimensional vector of ones by $\vone_d$, and use $\odot$ to denote the Hadamard product.

\paragraph{Overview.}
Consider a model $f$ with parameters $\vtheta \in \sR^d$ trained on a dataset  $\gD=\{\rvx_i, \rvy_i\}_{i=1}^N$. Suppose a subset of $\gD$, denoted as $\gD_f$, is requested for deletion. The remaining data is defined as $\gD_r \coloneqq \gD \setminus \gD_f$. Our objective is to develop the unlearning algorithm to generate a scrubbed model $f_u$, effectively removing the influence of $\gD_f$ from model $f$, while maintaining its utility over 
$\gD_r$.

In \algname, we initially identify critical parameters \wrt the forgetting data via connection sensitivity analysis~\cite{lee2018snip,lee2019signal}. We then reinitialize the top-$k$ percent of these key parameters, resulting in a trimmed model with parameters $\vtheta_t$. This step aims to diminish the model's memory of the forgetting data $\gD_f$. Nevertheless, this process runs the risk of erasing valuable information from the data set $\gD_r$ that we aim to retain.

To address this, the next phase of \algname concentrates on restoring the performance of the trimmed model on remaining data $\gD_r$, while ensuring that the influence of the forgetting data $\gD_f$ remains excluded. We accomplish this balance and obtain the final scrubbed model with parameters $\vtheta^\star$ through a gradient projection-based approach.

\subsection{Trimming}
\label{sec:Trimming}

To effectively identify salient parameters of a network with respect to $\gD_f$, one has to establish a good criterion to determine such salient connections. A commonly used criterion is the magnitude of the weights, with those exceeding a certain threshold deemed salient as suggested in~\cite{han2015learning_neurips}. Our requirement is more nuanced, as we need a measure that specifically discerns saliency based on $\gD_f$.
Therefore, we adopt a single-shot approach as proposed in~\cite{lee2018snip}, which we outline briefly to ensure clarity and completeness in our methodology.
Consider
\begin{align}
    \label{eq:cs_ob}
    \texttt{s}_j(\gD) \coloneqq \mathbb{E}_{\vx,\vy \sim \gD} \Big[\ell(\vtheta; \vx,\vy) - \ell((\vone_d - \ve_j) \odot \vtheta; \vx,\vy)\Big],
\end{align}
which measures the influence of parameter $j\in \{1, \ldots, d\}$ on a model in terms of the empirical risk for a given dataset $\gD$. Here, 
$\ell:\mathbb{R}^d\times \mathcal{X} \times \mathcal{Y} \to \mathbb{R}_+$ is the loss of the model and 
$\ve_j$ is the indicator vector of element $j$ (\ie, a binary vector with its $j$-th component set to one). 
Note that computing $\texttt{s}_j$ for each $j\in \{1, \ldots, d\}$ is prohibitively expensive, as it demands $d + 1$ forward passes through the dataset. 
An approximation to \Cref{eq:cs_ob}~\cite{han2015learning_neurips} is in the form of 
\begin{align}
    \label{eq:cs_final}
    \texttt{s}_j(\gD) \approx \mathbb{E}_{\vx,\vy \sim \gD} \Big[ 
    \frac{\partial \ell(\vtheta; \vx, \vy)}{\partial\vtheta_j} \vtheta_j\Big]\;.
\end{align}
Detailed proofs can be found in Appendix A.
\Cref{eq:cs_final} defines the sensitivity of parameter $j$ as the average (\ie, expectation) of a product, the gradient of the loss and the current value of the parameter. As such, $\texttt{s}_j$ encodes the loss sensitivity according to the parameter's magnitude. Intuitively, a parameter with a large value that significantly affects the loss, as captured by the gradient, is considered more influential or salient. The form in \Cref{eq:cs_final} offers several benefits, most importantly, the ability to calculate the sensitivity for all parameters with just a single sweep through the dataset.

To scrub the influence of the forgetting data $\gD_f$ in the model, we first obtain $\texttt{s}_j(\gD_f)$, saliency of parameters \wrt the forgetting data.
We then re-initialize the top-$k\%$ of the parameters based on their saliency rankings. This process is akin to performing a targeted `lobotomy' on the model specifically concerning the data $\gD_f$, thereby selectively erasing their influence. Unfortunately, this aggressive approach of reinitializing parameters can detrimentally impact the model's performance on the remaining dataset $\gD_r$, which we will correct in the next phase of \algname.

\begin{remark}
In principle, Scissorhands can benefit from any algorithms that can identify important connections \wrt $\mathcal{D}_f$.
The main reason behind choosing SNIP is its single-shot nature. As presented in the study~\cite{fan2023salun}, the gradient of loss \wrt the model parameters over the forgetting data can help identify important parameters \wrt $\mathcal{D}_f$ and hence help with erasing.
\end{remark}

\begin{remark}
In practice, we have observed that even a small subset of $\gD_f$ can be sufficient for the trimming.
For example, to unlearn a large set of data on CIFAR-100, our algorithm outperforms baselines by utilizing approximately 3\% of the forgetting data for trimming.
\end{remark}

\begin{remark}
Initializing weights of the trimmed neurons can take various forms. Empirically, we observed initializing with a uniform distribution is particularly effective. Details on the influence of different initialization strategies and the choice of 
k\% trimming on the model's performance will be discussed in \textsection~\ref{subsec:ablation}.
\end{remark}

\subsection{Repairing}
\label{sec:Repairing}
Following the parameter-trimming process, aimed at mitigating the influence of the forgetting data $\gD_f$, a challenge presents itself: the potential erasure of crucial information associated with the remaining data $\gD_r$.
A straightforward solution to recover the model utility is to relearn the model using the remaining data. However, this approach risks biasing the model towards features of the remaining data and inadvertently reintegrating information about the forgetting samples.
In the quest to ensure that models retain essential information from the remaining data $\gD_r$ while effectively forgetting $\gD_f$, our strategy involves maximizing the loss over the forgetting data $\gD_f$ while concurrently minimizing the loss over the remaining data $\gD_r$.
Therefore, to achieve the balance between these goals, we propose the objective of an efficient practice for unlearning:
\begin{align}
    \label{eq:obj}
    L(\vtheta, \gD_r, \gD_f) \coloneqq \mathcal{L}(\vtheta;\gD_r) - \lambda \mathcal{L}(\vtheta;\gD_f)\;.
\end{align}
\begin{algorithm}[tb]
\caption{The procedure of \algname.}
\label{alg:unl}
% \vspace{-\baselineskip}
\begin{multicols}{2}
\begin{algorithmic}[1]
\Procedure{Trimming}{$\vtheta, \gD_f$}
    \State $\mX, \vy \gets \{\rvx_i, \rvy_i \sim \gD_f|i \in [B]\}$.
    \State // Compute connection sensitivity
    \State Get $\texttt{s}_j(\vtheta; \mX, \vy), \hspace{0.2em} \forall j \in [1, d]$
    \Statex \hspace{2em} using \cref{eq:cs_final}.
    \State // Re-initialization
    \State Re-initialize parameters according
    \Statex \hspace{2em} to the top-$k$\% value of $\texttt{s}_j$.
    \State \Return $\vtheta_t$
\EndProcedure
\Statex
\Procedure{Repairing}{$\vtheta_t, \gD_f, \gD_r$}
    \State $\vtheta^{0} = \vtheta_t$.
    \For{$e=0$ to $E-1$}
        \State Get $L(\vtheta, \gD_r, \gD_f)$ (\cf \cref{eq:obj}).
        \State Get $\vg_o$ and $\vg_f$.
        \State // Get the optimal direction
        \If{$\langle \vg_o, \vg_f \rangle > 0$}
        \State Compute $v^{*}$ (\cf \cref{eq:solution}).
        \State $\vg = \vg_o - v^{*} \vg_f$.
        \State $\vg_o \gets \vg$.
        \EndIf
        \State $\vtheta^{e+1} \gets \vtheta^{e} - \eta \vg_o$.
    \EndFor
    \State \Return $\vtheta^*=\vtheta^E$.
\EndProcedure
\end{algorithmic}
\end{multicols}
\end{algorithm}

We aim to optimize \Cref{eq:obj} \wrt $\vtheta$ through gradient descent, ensuring minimal reintroduction of information about $\gD_f$. This is particularly challenging when there are similarities between samples in $\gD_f$ and $\gD_r$. To achieve this desideratum, it is vital that updates to the model do not improve (\ie, reduce) $\mathcal{L}(\vtheta; \gD_f)$.
To this end, we define $\vg_f = \nabla_{\vtheta} \mathcal{L}(\vtheta; \gD_f)$ as the gradient direction for $\gD_f$ and $\vg_o = \nabla_{\vtheta} L(\vtheta, \gD_r, \gD_f)$ as the gradient direction for optimizing \Cref{eq:obj}. The optimal descent direction $\vg \in \R^d$ should thus satisfy two criteria: \textbf{1.} It should exhibit maximum similarity to $\vg_o$ to ensure swift convergence. The notion of similarity can be captured by $\vg_o$ using $\left\| \vg - {\vg_o} \right\|^{2}_{2}$.
\textbf{2.} It should not align with $\vg_f$ to prevent improving $\mathcal{L}(\vtheta; \gD_f)$. This can be mathematically formalized as $\langle \vg_f, \vg \rangle \leq 0$. Consequently, after computing $\vg_o = \nabla_{\vtheta} L(\vtheta, \gD_r, \gD_f)$ and $\vg_f = \nabla_{\vtheta} \mathcal{L}(\vtheta; \gD_f)$ in each iteration, we formulate the following optimization challenge to identify the model update direction:
\begin{align}
\label{eq:proj}
    \argmin_{{\vg}} \quad &\frac{1}{2} \left\| \vg - {\vg_o} \right\|^{2}_{2}, \notag \\
    \text{s.t.} \quad &\langle \vg_f, \vg \rangle \leq 0.
\end{align}
This problem may be addressed using the Frank-Wolfe algorithm. Given the high dimensionality of gradients in neural networks, directly addressing the constraint optimization in \Cref{eq:proj} could become overwhelming. Inspired by~\cite{lopez2017gradient}, we propose to make use of the dual formulation of the problem \Cref{eq:proj} as:
\begin{align}
    \label{eq:solution}
    \sup_{v} \quad &\frac{1}{2} \vg_f^\top \vg_f v^2 - \vg_o^\top \vg_f v, \notag \\
    \text{s.t.} \quad &v \geq 0.
\end{align}
Then, the optimal descent direction $\vg$ is given from the solution $v^{*}$ in \Cref{eq:solution} as $\vg = \vg_o - v^{*}\vg_f $.
Detailed proofs can be found in Appendix A. By doing so, we ensure a balance between unlearning and retaining utility.

\begin{remark}
The primal problem (\cf \Cref{eq:proj}) involves optimizing over the high-dimensional space of gradient vectors, while the dual problem (\cf \Cref{eq:solution}) simplifies this problem to optimize over a single scalar variable $v \in \R$. This dimensionality reduction can significantly decrease the computational complexity, especially for large-scale neural networks.
\end{remark}

\subsection{Algorithm Description}

\Cref{alg:unl} describes the procedure of \algname in detail.
We first identify the important parameters \wrt the forgetting data in single-shot via connection sensitivity in networks, then re-initialize these parameters for erasing the influence of the forgetting data $\gD_f$ in the given model. To repair the performance of the trimmed model on the remaining data $\gD_r$, while ensuring the exclusion of information associated with the forgetting data $\gD_f$, we employ the gradient projection-based optimization algorithm.
By doing so, we can scrub the influence of the forgetting data while retaining model utility on the remaining data.
The loss depicted in \Cref{eq:obj} can be viewed as a scalarization of a Multi-Objective Optimization (MOO) problem, namely minimizing $\big(\mathcal{L}(\theta;\mathcal{D}_r), - \mathcal{L}(\theta;\mathcal{D}_f)\big)^\top$. A common issue in MOO is the gradient conflict, as optimizing one objective could hinder another one. Our proposal on gradient projection is designed to address and mitigate the gradient conflict. Empirically (see \Cref{tab:ablation}), we studied the effect of removing gradient projection.

\section{Related Work}
\label{sec:relatedwork}

The development of efficient machine unlearning methods~\cite{romero2007incremental, karasuyama2010multiple, cao2015towards, ginart2019making, bourtoule2021machine, peste2021ssse, wu2020deltagrad, guo2019certified, golatkar2020eternal, mehta2022deep, sekhari2021remember,jia2023model,tarun2023deep,nguyen2022survey,goel2022towards,fan2024challenging,zhang2024unlearncanvas} has gained prominence. Applications of machine unlearning span various domains, including regression tasks~\cite{tarun2023deep}, federated learning~\cite{liu2022right,liu2021federaser,halimi2022federated, wu2022federated,wang2022federated}, graph neural networks~\cite{chen2022graph,cheng2023gnndelete}, and diffusion models~\cite{gandikota2023erasing,zhang2023forget,heng2023selective,gandikota2023unified,fan2023salun,wu2024erasediff,zhang2024defensive}, as well as scenarios where training data are not available~\cite{tarun2023fast,chundawat2023zero}.

Retraining the model from scratch without forgetting data is typically considered the standard gold unlearning algorithm~\cite{thudi2022unrolling,fan2023salun}. However, this is often deemed impractical as most in-production models require extensive training periods and considerable computing resources. While fine-tuning the model for a new task might lead to forgetting previous knowledge (\ie, catastrophic forgetting)~\cite{lopez2017gradient}, it fails to adequately erase the data influence in the models.
Approximate unlearning methods, as such, become attractive alternatives. We will briefly discuss these methods in image classification and image generation.

\noindent
\textbf{Unlearning in Classification Models.}
Most unlearning algorithms are based on the influence function~\cite{sekhari2021remember, guo2019certified, neel2021descent} and the Fisher Information Matrix (FIM)~\cite{golatkar2020eternal,golatkar2020forgetting,golatkar2021mixed}.
Influence function-based methods estimate the influence of the particular training data points on the model's predictions, and Fisher unlearning methods assume that the unlearned model and the retrained model are close to each other, Golatkar \etal~\cite{golatkar2020eternal} leverage FIM and hide the difference between the unlearned models and retrained models via adding noise. Golatkar \etal~\cite{golatkar2020eternal} also present an upper bound for the amount of retained information, offering a quantifiable measure of the effectiveness of the unlearning process.
Such approximate unlearning methods need to compute the Hessian matrix, or FIM \wrt the data.
This may render them impractical for certain scenarios, such as federated learning, where computational resources are distributed and limited.
To mitigate these computational demands, Mehta \etal~\cite{mehta2022deep} introduce L-CODEC, a strategy aimed at pre-selecting a subset of parameters, allowing for more efficient computation.
Jia \etal~\cite{jia2023model} recently propose to fuse the model sparsity into the unlearning algorithm, helping improve approximate unlearning methods, and achieving effective and efficient unlearning effects.

\noindent
\textbf{Concept Erasure in Generative Models.}
The advent of generative models, particularly those converting text to images, has been a significant milestone in the field of artificial intelligence. A notable concern is the risk of it being tainted or manipulated~\cite{chen2023trojdiff,rando2022red}, resulting in not-safe-for-work (NSFW) generations~\cite{schramowski2023safe}, as these models leverage training data from a wide array of open sources.
To address these challenges, data censoring~\cite{gandhi2020scalable,birhane2021large,nichol2021glide,schramowski2022can} to exclude black-listed images is employed. Studies~\cite{gandikota2023erasing, schramowski2023safe} introduce methods to update the models away from the inappropriate concepts. Heng and Soh~\cite{heng2023selective} recently propose to adopt Elastic Weight Consolidation (EWC) and Generative Replay (GR) to effectively unlearn without access to the training data for a wide range of generative models.

\noindent
\textbf{Unlearning across Domains.}
Recently, Fan \etal~\cite{fan2023salun} highlight the difficulty in cross-domain applicability of the machine unlearning algorithms, \ie, existing machine unlearning methods designed for classification tasks cannot be effective when applied to image generation.
To address this gap, they introduce a novel method, Saliency unlearning (SalUn), which shifts attention from the entire model to target parameters. SalUn achieves state-of-the-art performance and is effective in both image classification and image generation tasks.

In this work, we propose \algname which could achieve improved performance compared to existing methods across various scenarios.
Inspired by~\cite{jia2023model} that apply $\ell_1$ regularization to simply fuse sparsity into the unlearning methods, we identify and re-initialize the key model parameters \wrt the forgetting data via connection sensitivity.
Concurrently, Fan \etal~\cite{fan2023salun} also explore unlearning via salience scores, determining weight salience via the gradient of the forgetting loss \wrt model parameters, and employing random labeling for model fine-tuning.
While there are similarities in our overarching goals, the methodologies present unique perspectives and solutions to the problem of unlearning.
Fan \etal~\cite{fan2023salun} only focus on the salient weights, while our mechanism aims to first `destroy' the key information and then relearn for the whole model.
Additionally, using the forgetting data with random labels could potentially lead to memorizing information about the forgetting data points~\cite{zhang2021understanding}.
In contrast, \algname excludes knowledge about the forgetting data via a gradient projection-based approach, resulting in the effective erasure of information about the forgetting data while preserving the knowledge about the retained data.

\section{Experimental Evaluation}
\label{sec:experiment}

In this section, we illustrate how \algname effectively eliminates the data influence in the models.
For sample-wise forgetting, where the forgetting data shares the same distribution as the training data, we evaluate \algname on SVHN~\cite{netzer2011reading}, CIFAR-10 and CIFAR-100~\cite{krizhevsky2009learning}.
We further extend our evaluation on CelebAMask-HQ~\cite{lee2020maskgan} where targeting to remove the entirety of specific identities.
Additionally, our experimentation encompasses the open-source text-to-image model, Stable Diffusion v1.4~\cite{rombach2022high}, which is conditioned on CLIP text embeddings through the cross-attention mechanism.
Further detailed experimental setups and additional results can be found in Appendix B.

\noindent
\textbf{Baselines.}
We primarily compare against the following standard baselines that are frequently employed in machine unlearning, alongside recently proposed SOTA methods:
(\romannumeral1) \textbf{Retrain}: models obtained by only retraining from scratch on $\gD_r$. This will provide the performance of the oracle for us.
(\romannumeral2) \textbf{Fine-tuning (FT)}~\cite{warnecke2021machine}: models that are fine-tuned on $\gD_r$, \ie, taking advantage from catastrophic forgetting in neural networks to unlearn.
(\romannumeral3) \textbf{Gradient ascent (GA)}~\cite{thudi2022unrolling}: gradient ascent on $\gD_f$. This will provide a simple way of unlearning by making the performance of the model worse on $\gD_f$.
(\romannumeral4) \textbf{Influence unlearning (IU)}~\cite{koh2017understanding}: utilizes the influence function method to estimate the change in model parameters when transitioning from the unscrubbed model to the retrained model.
(\romannumeral5) \textbf{Boundary shrink (BS)}~\cite{chen2023boundary} and (\romannumeral6) \textbf{Boundary expanding (BE)}~\cite{chen2023boundary}: shift the decision boundary of the original model to imitate that of the retrained model.
(\romannumeral7) \textbf{$\ell_1$-sparse}~\cite{jia2023model}: fine-tuning models on $\gD_r$ with $\ell_1$-norm sparse regularization.
(\romannumeral8) \textbf{Saliency unlearning (SalUn)}~\cite{fan2023salun}: adopt weight saliency and random labeling for unlearning.

\noindent
\textbf{Metrics.}
To assess the unlearning algorithms, we employ several metrics:
(\romannumeral1) Accuracy: the accuracy of the model on $\gD_f$ (denoted as \textbf{$\text{Acc}_{\gD_f}$}), $\gD_r$ (denoted as \textbf{$\text{Acc}_{\gD_r}$}) and the test set (denoted as \textbf{$\text{Acc}_{\gD_t}$}). It provides insight into how well the model performs after undergoing the unlearning process.
(\romannumeral2) \textbf{Membership inference attack (MIA)}: a standard metric for verifying the unlearning effect.
A classifier $\phi$ is trained using both training data (marked with a label of 1, indicating data points included in training) and test data (marked with a label of 0, indicating data points not seen during training), and subsequently assessed on $\gD_f$.
An effective unlearning method should make it challenging to infer whether a particular sample was part of the training set, mitigating potential privacy breaches.
In this context, we define MIA as $P_{\phi}(\vy=0|\vx_f)$, representing the probability assigned by the classifier $\phi$ that a given sample $\vx_f$ was not included in the training data.
(\romannumeral3)\textbf{Avg. Gap}~\cite{fan2023salun}: the average of the performance gaps measured in accuracy-related metrics: Avg. Gap $=(|\text{Acc}_{\gD_t} - \text{Acc}^\ast_{\gD_t}| + |\text{Acc}_{\gD_f} - \text{Acc}^\ast_{\gD_f}| + |\text{Acc}_{\gD_r} - \text{Acc}^\ast_{\gD_r}| + |\text{MIA} - \text{MIA}^\ast|)/4$.
where $\text{Acc}^\ast_{\gD_t}, \text{Acc}^\ast_{\gD_f}, \text{Acc}^\ast_{\gD_r}$ and $\text{MIA}^\ast$ are metric values of the retrained model.
A better performance of an unlearning method corresponds to a lower performance gap with retraining, measuring how close the unlearned model is to the retrained model.
(\romannumeral4) \textbf{Run-time efficiency (RTE)}~\cite{jia2023model}: evaluate the computational efficiency of the unlearning method. 
Specifically, we have $\text{RTE} = {\text{T}}/{\text{T}_r}$, where $\text{T}$ and $\text{T}_r$ present the time for the unlearning methods to get the scrubbed models and the time for retraining from scratch with $\gD_r$, respectively.
An efficient unlearning method should aim for minimal computational overhead compared to the baseline.
(\romannumeral5) \textbf{Relearn time}~\cite{golatkar2020eternal,chundawat2023zero,tarun2023deep}: the number of epochs to fine-tune the scrubbed models for regaining performance on $\gD_f$ (\ie, reach the original models' accuracy on $\gD_f$).

\subsection{Results on Classification Task}
\begin{table}[tb]
    \centering
    \caption{Quantitative results for forgetting 10\% data.}
    \label{tab:sample_level}
    \begin{tabular}{llccccc}
        \toprule
         &Method &$\text{Acc}_{\gD_f}(\downarrow)$ &$\text{Acc}_{\gD_t}(\uparrow)$ &$\text{Acc}_{\gD_r}(\uparrow)$ &MIA$(\uparrow)$ &Avg. Gap \\
         \midrule
         \multirow{9}{*}{CIFAR-100}
         &Retrain &75.13\scriptsize{$\pm$0.85} &74.69\scriptsize{$\pm$0.08} &99.98\scriptsize{$\pm$0.01} &50.22\scriptsize{$\pm$0.62}  &- \\
         \cdashlinelr{2-7}
         &FT~\cite{warnecke2021machine} &97.98\scriptsize{$\pm$1.36} &75.28\scriptsize{$\pm$0.12}  &99.95\scriptsize{$\pm$0.02} &9.64\scriptsize{$\pm$3.6} &16.01 \\
         &GA~\cite{thudi2022unrolling}  &98.00\scriptsize{$\pm$1.34} &75.59\scriptsize{$\pm$0.11}  &98.24\scriptsize{$\pm$1.16} &5.00\scriptsize{$\pm$2.25} &17.68  \\
         &IU~\cite{koh2017understanding}&95.67\scriptsize{$\pm$4.82} &72.13\scriptsize{$\pm$4.58}  &96.14\scriptsize{$\pm$4.51} &9.43\scriptsize{$\pm$5.98} &16.93  \\
         &BE~\cite{chen2023boundary} &97.94\scriptsize{$\pm$1.38} &74.16\scriptsize{$\pm$0.09} &98.12\scriptsize{$\pm$1.24} &7.60\scriptsize{$\pm$3.05} &16.96  \\
         &BS~\cite{chen2023boundary} &97.65\scriptsize{$\pm$1.48} &73.20\scriptsize{$\pm$0.18} &97.93\scriptsize{$\pm$1.30} &8.24\scriptsize{$\pm$3.23} &17.01  \\
         &$\ell_1$-sparse~\cite{jia2023model} &96.35\scriptsize{$\pm$0.67} &70.06\scriptsize{$\pm$0.46} &96.35\scriptsize{$\pm$0.67} &21.33\scriptsize{$\pm$1.95} &14.59 \\
         &SalUn~\cite{fan2023salun} &88.56\scriptsize{$\pm$1.18} &71.34\scriptsize{$\pm$0.48} &99.40\scriptsize{$\pm$0.35} &74.66\scriptsize{$\pm$2.48} &10.45   \\
         &Ours &68.76\scriptsize{$\pm$1.81} &73.17\scriptsize{$\pm$0.24} &99.24\scriptsize{$\pm$0.30} &42.42\scriptsize{$\pm$2.06} &\textbf{4.11}  \\
         \midrule
         \multirow{9}{*}{CIFAR-10}
         &Retrain &94.81\scriptsize{$\pm$0.53} &94.26\scriptsize{$\pm$0.14} &100.0\scriptsize{$\pm$0.00} &13.05\scriptsize{$\pm$0.64} &- \\
         \cdashlinelr{2-7}
         &FT~\cite{warnecke2021machine} &99.15\scriptsize{$\pm$0.46} &93.83\scriptsize{$\pm$0.45} &99.84\scriptsize{$\pm$0.11} &3.01\scriptsize{$\pm$0.93} &3.74 \\
         &GA~\cite{thudi2022unrolling}  &99.66\scriptsize{$\pm$0.23} &94.57\scriptsize{$\pm$0.01} &99.62\scriptsize{$\pm$0.25} &0.91\scriptsize{$\pm$0.29}  &4.42 \\
         &IU~\cite{koh2017understanding}&98.08\scriptsize{$\pm$2.10} &91.91\scriptsize{$\pm$2.73} &98.01\scriptsize{$\pm$2.26} &4.01\scriptsize{$\pm$3.44} &4.16   \\
         &BE~\cite{chen2023boundary} &99.41\scriptsize{$\pm$0.38} &93.79\scriptsize{$\pm$0.15} &99.41\scriptsize{$\pm$0.38} &16.16\scriptsize{$\pm$0.78} &2.19  \\
         &BS~\cite{chen2023boundary} &99.60\scriptsize{$\pm$0.25} &94.24\scriptsize{$\pm$0.07} &99.56\scriptsize{$\pm$0.54} &4.46\scriptsize{$\pm$0.33} &3.46 \\
         &$\ell_1$-sparse~\cite{jia2023model} &94.17\scriptsize{$\pm$0.49} &90.64\scriptsize{$\pm$0.52} &96.64\scriptsize{$\pm$0.54} &11.87\scriptsize{$\pm$0.61} &2.20   \\
         &SalUn~\cite{fan2023salun} &98.07\scriptsize{$\pm$0.42} &93.92\scriptsize{$\pm$0.25} &99.89\scriptsize{$\pm$0.07} &17.93\scriptsize{$\pm$0.37} &2.15   \\
         &Ours &95.40\scriptsize{$\pm$1.48} &92.92\scriptsize{$\pm$0.48} &98.93\scriptsize{$\pm$0.57} &9.56\scriptsize{$\pm$2.13} &\textbf{1.62}  \\
         \midrule
         \multirow{9}{*}{SVHN}
         &Retrain &91.81\scriptsize{$\pm$1.11} &91.17\scriptsize{$\pm$1.77} &97.73\scriptsize{$\pm$1.12}  &15.74\scriptsize{$\pm$1.28} &- \\
         \cdashlinelr{2-7}
         &FT~\cite{warnecke2021machine}  &99.60\scriptsize{$\pm$0.24} &95.28\scriptsize{$\pm$0.04} &99.99\scriptsize{$\pm$0.00} &3.85\scriptsize{$\pm$0.4} &6.51  \\
         &GA~\cite{thudi2022unrolling}   &98.41\scriptsize{$\pm$0.23} &92.87\scriptsize{$\pm$0.06} &98.52\scriptsize{$\pm$0.29} &5.96\scriptsize{$\pm$0.38} &4.72 \\
         &IU~\cite{koh2017understanding} &92.47\scriptsize{$\pm$1.62} &86.92\scriptsize{$\pm$2.06} &93.36\scriptsize{$\pm$1.82} &17.00\scriptsize{$\pm$2.67} &2.63  \\
         &BE~\cite{chen2023boundary}  &98.48\scriptsize{$\pm$0.33} &92.62\scriptsize{$\pm$0.05} &98.44\scriptsize{$\pm$0.29} &6.77\scriptsize{$\pm$0.36} &4.45  \\
         &BS~\cite{chen2023boundary}  &98.29\scriptsize{$\pm$0.32} &92.48\scriptsize{$\pm$0.02} &98.36\scriptsize{$\pm$0.29} &6.74\scriptsize{$\pm$0.39}  &4.36 \\
         &$\ell_1$-sparse~\cite{jia2023model} &97.79\scriptsize{$\pm$0.10} &93.59\scriptsize{$\pm$3.18} &99.49\scriptsize{$\pm$0.23} &7.57\scriptsize{$\pm$0.29} &4.65  \\
         &SalUn~\cite{fan2023salun} &96.29\scriptsize{$\pm$4.80} &93.59\scriptsize{$\pm$3.18} &97.41\scriptsize{$\pm$4.96} &24.47\scriptsize{$\pm$4.08} &3.99  \\
         &Ours &91.62\scriptsize{$\pm$1.18} &93.05\scriptsize{$\pm$0.56} &99.51\scriptsize{$\pm$0.14} &17.62\scriptsize{$\pm$2.54} &\textbf{1.43} \\
        \bottomrule
    \end{tabular}
\end{table}
\begin{table}[tb]
    \centering
    \caption{Quantitative results for forgetting 10\% identities on the CelebAMask-HQ.}
    \label{tab:class_level}
    \begin{tabular}{llccccccccc}
        \toprule
         &Method &$\text{Acc}_{\gD_f}(\downarrow)$& &$\text{Acc}_{\gD_t}(\uparrow)$& &$\text{Acc}_{\gD_r}(\uparrow)$& &MIA$(\uparrow)$& &Avg. Gap \\
         \midrule
         &Retrain &0.00\scriptsize{$\pm$0.00}& &87.02\scriptsize{$\pm$0.80}& &99.96\scriptsize{$\pm$0.01}& &100.0\scriptsize{$\pm$0.00}& &- \\
         \cdashlinelr{2-11}
         &FT~\cite{warnecke2021machine} &99.94\scriptsize{$\pm$0.12}& &88.59\scriptsize{$\pm$0.59}& &99.97\scriptsize{$\pm$7.02}& &5.28\scriptsize{$\pm$2.03}& &49.06   \\
         &GA~\cite{thudi2022unrolling}  &87.60\scriptsize{$\pm$8.71}& &81.22\scriptsize{$\pm$2.11}& &99.74\scriptsize{$\pm$0.26}& &51.37\scriptsize{$\pm$5.96}& &35.56  \\
         &IU~\cite{koh2017understanding} &88.92\scriptsize{$\pm$10.25}& &70.24\scriptsize{$\pm$11.77}& &95.27\scriptsize{$\pm$5.07}& &29.59\scriptsize{$\pm$18.59}& &45.20 \\
         &BE~\cite{chen2023boundary} &69.07\scriptsize{$\pm$2.73}& &44.11\scriptsize{$\pm$2.08}& &95.58\scriptsize{$\pm$1.23}& &46.24\scriptsize{$\pm$5.90}& &42.53 \\
         &BS~\cite{chen2023boundary} &98.18\scriptsize{$\pm$1.92}& &81.92\scriptsize{$\pm$0.27}& &99.86\scriptsize{$\pm$0.03}& &45.93\scriptsize{$\pm$5.11}& &39.36   \\
         &$\ell_1$-sparse~\cite{jia2023model} &98.81\scriptsize{$\pm$0.72}& &89.37\scriptsize{$\pm$0.70}& &99.97\scriptsize{$\pm$0.00}& &76.78\scriptsize{$\pm$5.66}& &31.10 \\
         &SalUn~\cite{fan2023salun} &0.00\scriptsize{$\pm$0.00}& &78.36\scriptsize{$\pm$1.34}& &96.90\scriptsize{$\pm$1.11}& &100.0\scriptsize{$\pm$0.00}& &2.93 \\
         &Ours &1.52\scriptsize{$\pm$2.73}& &80.18\scriptsize{$\pm$6.60}& &97.20\scriptsize{$\pm$3.81}& &99.83\scriptsize{$\pm$0.35}& &\textbf{2.82}  \\
        \bottomrule
    \end{tabular}
\end{table}
\begin{figure}[tb]
  \centering
  \includegraphics[width=0.86\textwidth, keepaspectratio=True]{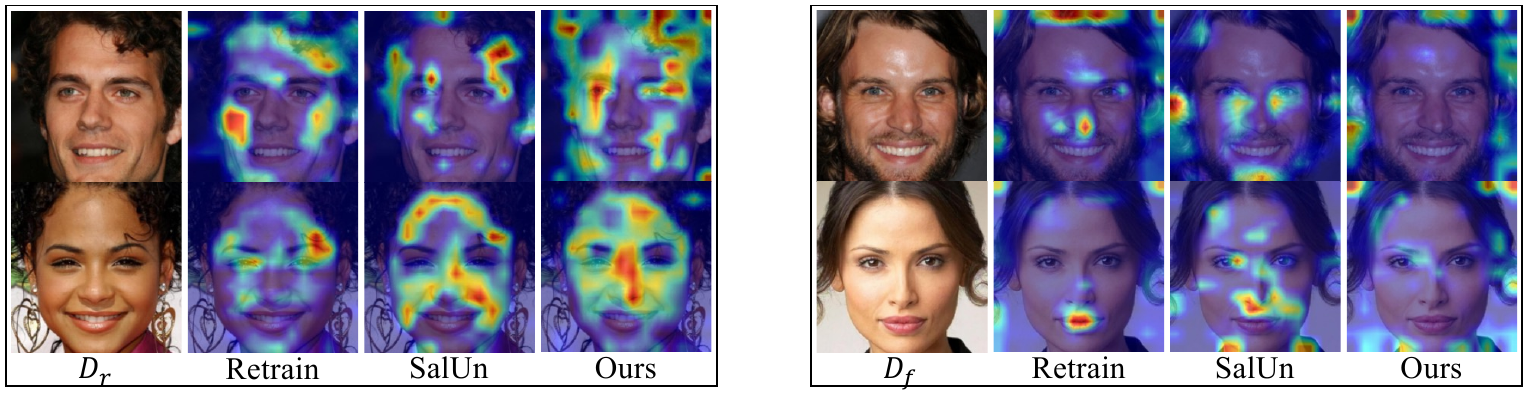}
  \caption{Visualizations of regions where models focus on generated by GradCAM~\cite{selvaraju2017grad}.
  }
  \label{fig:gradcam}
\end{figure}

In this experiment on SVHN, CIFAR-10, and CIFAR-100, we try to forget randomly selected 10\% of the training data; on CelebAMask-HQ, we attempt to forget randomly selected 10\% identities among 307 identities. We further apply GradCAM~\cite{selvaraju2017grad} to visualize regions where models focus on \with and \without machine unlearning algorithms. In brief, the results suggest that \algname has successfully induced forgetting for the relevant samples and classes, with minor degradation in model performance over the remaining data and classes.
We will discuss the results in depth below.

\paragraph{Sample-wise unlearning.}
\Cref{tab:sample_level} presents the results when forgetting randomly selected samples.
\algname achieves the lowest average performance gap on all the presented datasets and shows good generalization ability compared to Retrain. For example, on SVHN, \algname achieves $\sim$ 91.62\% accuracy on the forgetting data $\gD_f$, closely matching the 91.81\% accuracy of a fully retrained model.
Furthermore, the retrained model reaches around 91\% accuracy on the test dataset, \algname exhibits an accuracy of roughly 93\%, showcasing the enhanced generalization of our scrubbed model.

When evaluating solely based on forgetting accuracy on CIFAR-10, the $\ell_1$-sparse method might seem to be the most effective baseline, achieving approximately 94.17\%. Nonetheless, this perceived advantage is offset by a decline in both test accuracy and the accuracy of retained data. In contrast, \algname shows a good trade-off between preserving model utility on the remaining data, generalization on unseen sets, and erasing the influence of the forgetting data.

Similarly, on CIFAR-100, SalUn may initially stand out with its impressive MIA accuracy of approximately 74.66\%, relying solely on a single metric to gauge the performance of machine unlearning can be misleading, as it might not fully capture the method's effectiveness~\cite{fan2023salun}. While SalUn achieves a forgetting accuracy of about 88.56\% and a test accuracy of 71.34\%, \algname demonstrates a slightly lower forgetting accuracy of around 68.76\% but achieves a higher test accuracy of 73.17\%. This indicates that \algname achieves the best trade-off between eradicating data traces and generalizing to unseen data.
Soon, we will show that on this dataset, the relearning time of \algname is greater than 200 to regain the performance on $\gD_f$, which indicates that \algname has efficiently scrubbed information about $\gD_f$.

Overall, \algname illustrates its capability to adeptly balance the objectives of data removal, model utility preservation, and generalization to unseen data, ensuring that the influence of forgotten data is minimized without significantly compromising the overall performance and utility of the model.

\paragraph{Class-wise unlearning.}
Furthermore, \Cref{tab:class_level} presents the outcomes of our analysis for the scenario where forgetting 10\% of identities on the CelebAMask-HQ. Images are rescaled to $224 \times 224$, and a model pre-trained with ImageNet1K~\cite{deng2009imagenet} is employed. In this context, \algname manifests the smallest average performance gap with retrained models.
Notably, the Retrain method achieves perfect metrics in terms of MIA accuracy, which is 100\%, and completely erases the identity information from the dataset, as indicated by a 0.00\% accuracy on the forgotten data $\gD_f$. However, retraining proves impractical and limited for unlearning purposes in general, functioning merely as an oracle in our experiments.

Among the baselines, fine-tuning (FT) and $\ell_1$-sparse exhibit higher accuracies on the remaining data $\gD_f$ and unseen data $\gD_t$; yet, their accuracies on forgetting remain high as well, highlighting a less effective balance between eradicating data traces and generalizing to unseen data.
In contrast, \algname and SalUn demonstrate superior capabilities in effectively forgetting identities with minimal impact on the models' overall performance. SalUn achieves an MIA accuracy of 100\% and a forgetting accuracy of 0.00\%, effectively nullifying the identity information similar to the retrained models. \algname slightly surpasses SalUn in terms of the average performance gap with retrained models, maintaining a high MIA accuracy of 99.83\% and a significantly low accuracy on the forgotten data $\gD_f$ (\ie, 1.52\%), while achieving a test accuracy of 80.18\% and an accuracy of 97.20\% on the retained data.

We further employ the GradCAM~\cite{selvaraju2017grad} to illustrate the focus area of models \with and \without machine unlearning algorithms. As shown in \Cref{fig:gradcam}, when evaluated on the remaining data $\gD_r$, both SalUn and \algname even pay more attention to the facial features for identity classification than the retrained model. Conversely, when evaluated on the forgetting data $\gD_f$, the scrubbed model by \algname shifts its attention towards regions that are least associated with the facial features.
We hypothesize that in the scenario where the task involves forgetting identities, both the remaining data and the forgetting data are face images, the scrubbed models would need a more nuanced approach to distinguish between individuals.
Specifically, because the task at hand requires the discernment of subtle differences across facial features to effectively differentiate and forget specific identities while retaining others, the models adapt by developing a refined sensitivity towards those facial characteristics that are most indicative of individual identity. This adaptation enables the models to maintain high accuracy on the remaining data by focusing more intently on the facial features that are crucial for identity recognition, thereby enhancing their ability to generalize and accurately classify identities.

These results underscore our proposed machine unlearning algorithm \algname's superior capabilities in balancing identity forgetting and model utility performance, \algname not only minimizes privacy risks but also maintains the integrity and applicability of the model to unseen data.

\subsection{Ablation Study}
\label{subsec:ablation}
\begin{table}[tb]
    \centering
    \caption{Influence of projection and initialization strategies (re-initializing parameters with uniform distribution $\gU$, Gaussian distribution $\gN$, and constant value of $\vzero$/$\vone$).}
    \label{tab:ablation}
    \begin{tabular}{cc|ccccc}
        \hline
        Projection &Initialization &$\text{Acc}_{\gD_f}(\downarrow)$ &$\text{Acc}_{\gD_t}(\uparrow)$ &$\text{Acc}_{\gD_r}(\uparrow)$ &MIA$(\uparrow)$ &Avg. Gap \\
         \hline
        \cmark &$\gU$ &95.40\scriptsize{$\pm$1.48} &92.92\scriptsize{$\pm$0.48} &98.93\scriptsize{$\pm$0.57} &9.56\scriptsize{$\pm$2.13} &1.62  \\
        \xmark &$\gU$ &98.14\scriptsize{$\pm$0.60} &93.14\scriptsize{$\pm$0.16} &99.73\scriptsize{$\pm$0.16} &6.08\scriptsize{$\pm$0.78} &2.92  \\
        \cmark &$\gN$ &94.68\scriptsize{$\pm$2.50} &92.13\scriptsize{$\pm$2.20} &97.94\scriptsize{$\pm$2.71} &10.18\scriptsize{$\pm$3.81} &1.80  \\
        \cmark &$\vzero$ &96.10\scriptsize{$\pm$2.05} &92.92\scriptsize{$\pm$0.99} &99.09\scriptsize{$\pm$1.19} &8.76\scriptsize{$\pm$3.30} &1.96  \\
        \xmark &$\vone$ &84.28\scriptsize{$\pm$0.49} &83.38\scriptsize{$\pm$0.58} &87.18\scriptsize{$\pm$0.90} &20.67\scriptsize{$\pm$0.41} &10.46  \\
        \hline
    \end{tabular}
\end{table}
\begin{figure}[tb]
  \centering
  \includegraphics[width=0.99\textwidth, keepaspectratio=True]{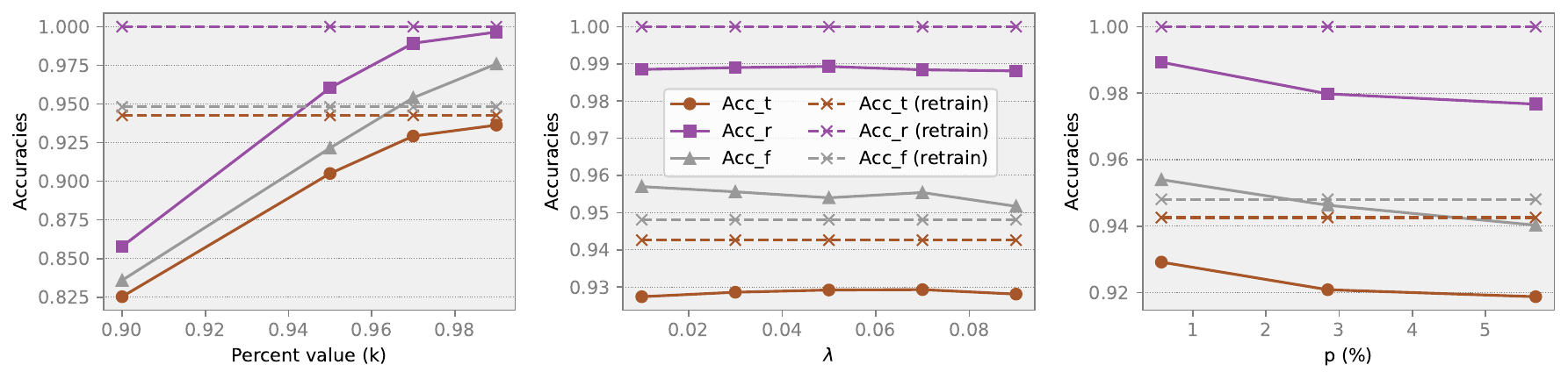}
  \caption{Influence of the percent value $k$ in the trimming process, the balance term $\lambda$ of \Cref{eq:obj}, and the ratio of forgetting data used in the trimming process.}
  \label{fig:ablation}
\end{figure}

\paragraph{\textbf{Component analysis.}}
As detailed in \Cref{tab:ablation}, the inclusion of different procedures in \algname has a noticeable impact.
Notably, the employment of the projection mechanism plays a crucial role in eliminating the influence of forgetting data $\gD_f$, affecting the overall utility of the model. The rationale behind the projection process is to preclude the reacquisition of critical information about the forgetting data $\gD_f$ during the repair phase.
Moreover, \Cref{tab:ablation} reveals that initializing parameters with a uniform distribution $\gU$ or opting for zero initialization facilitates effective unlearning, in contrast to re-initializing parameters with a constant value of one, which impedes finding a viable solution to \Cref{eq:solution}.

\paragraph{\textbf{Hyper-parameter impact.}}
\Cref{fig:ablation} presents the effects of varying hyper-parameters, such as the percentage value $k$\% for the trimming process, the balance term $\lambda$ of \Cref{eq:obj} in the repairing phase, and the ratio of forgetting data used in the trimming process.
An increase $k$ results in higher accuracies on data, as fewer parameters associated with the forgetting data $\gD_f$ are re-initialized.
Empirically, we observe choosing a value of $k\geq 0.9$ will result in stable performances. That said, this is a hyperparameter of the algorithm and can be chosen by measuring the performances on $D_f$ and $D_r$ during training.
The hyper-parameter $\lambda$ controls the balance between retaining and forgetting data, the accuracies on test data and remaining data show a peak at $\lambda=0.05$ when forgetting 10\% data on CIFAR-10. Generally, as $\lambda$ escalates, it places greater emphasis on the process of data omission, leading to a decrease in forgetting accuracy.
As the ratio $p$ increases, all accuracy metrics exhibit a decline, suggesting that employing more forgetting data for the purpose of forgetting in order to isolate critical knowledge about the forgetting data also inadvertently results in the loss of pertinent information regarding the remaining data. This phenomenon occurs particularly when the data designated for forgetting shares the same distribution as the data that is retained. In essence, while aiming to enhance the specificity of the forgetting process by pinpointing essential details about the data to be forgotten, there's a risk of simultaneously diminishing knowledge pertinent to the retained data.

\subsection{Case Study: Stable Diffusion}
\begin{figure}[tb]
  \centering
  \includegraphics[width=0.98\textwidth, keepaspectratio=True]{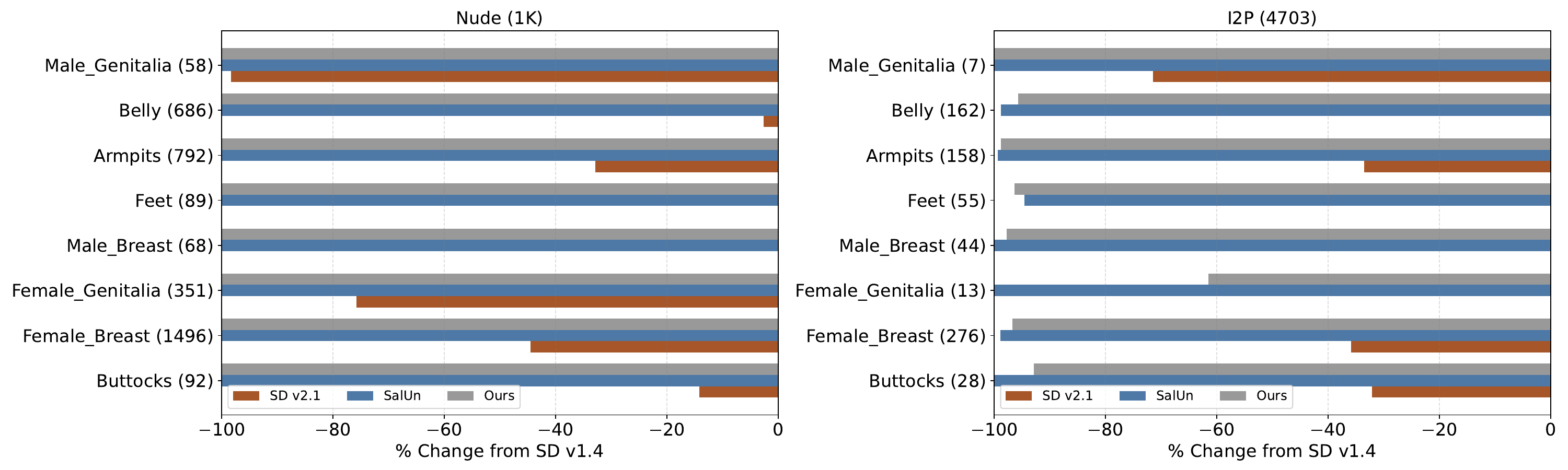}
  \caption{Quantity of nudity content detected using the NudeNet classifier from 1K sampled images and I2P data. We observed a high false positive rate for exposed female genitalia/breast using the NudeNet classifier on generated I2P images. The flagged images can be found in Appendix B.}
  \label{fig:i2p_nude_sd}
\end{figure}
\begin{figure}[tb]
  \centering
  \includegraphics[width=0.98\textwidth, keepaspectratio=True]{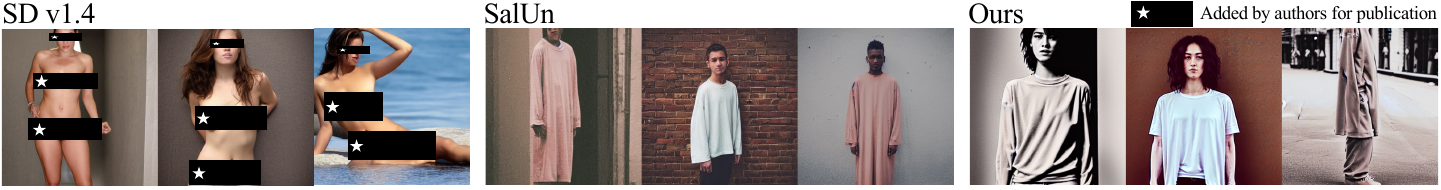}
  \caption{Sample images with the prompt from $c_f=$\{`nudity', `naked', `erotic', `sexual'\} generated by SDs \with and \without machine unlearning algorithms. Best viewed in color.}
  \label{fig:sd}
\end{figure}

We also employed our algorithm to mitigate the generation of inappropriate content in Stable Diffusion (SD) models~\cite{rombach2022high}.
Specifically, our primary objective is to effectively eliminate the concept of nudity from the model's generative capabilities. 
To this end, followed~\cite{heng2023selective} and employed SD v1.4 for sampling with 50 time steps.
We evaluate on 1K generated images with prompts $c_f=$\{`nudity', `naked', `erotic', `sexual'\} and 4703 generated images with I2P~\cite{schramowski2023safe} using the open-source NudeNet classifier.

\Cref{fig:i2p_nude_sd,fig:sd} present the results for SD v2.1 (trained on a dataset filtered for nudity), SalUn, and \algname.
In \Cref{fig:i2p_nude_sd}, y-axis denotes the number of exposed body parts generated by the SD v1.4 model, it presents the percentage change in exposed body parts \wrt SD v1.4.
When comparing on I2P data, our algorithms outperform SalUn on `Feet', while coming short on `Female Breast and Genitalia'. While this could suggest the superiority of SalUn on this data, the gap is not that big in reality, as out of the 9 images flagged for containing exposed female breasts, 7 were inaccurately identified, and none of the flagged images depicted exposed female genitalia. 
The use of the NudeNet classifier on generated I2P images exhibits a significant rate of false positives, as highlighted in~\cite{heng2023selective} as well.
When we use prompts $c_f$ explicitly associated with nudity, no exposed sensitive content is detected for both \algname and SalUn methods.
We can conclude that, \algname significantly reduces the amount of nudity content compared to SD v1.4, SD v2.1, showing comparable results to SalUn.

\section{Limitations and Broader Impact}

Our proposed unlearning algorithm, \algname, can achieve comparable performance to state-of-the-art unlearning methods.
We addressed potential issues of conflicting gradients arising from direct conflicts between features or patterns in the forgetting and remaining data, ensuring training stability.
However, the limitation is the incomplete concept erasure shown in Figure 8 in Appendix B, and the potential for biased models due to the selective deletion of information. This is a crucial consideration, especially when unlearning algorithms are applied with malicious intent. Furthermore, in experiments on stable diffusion, while our method can successfully erase specific concepts, finding an optimal balance between forgetting and maintaining high image quality remains challenging. There is also a risk that unlearning algorithms might be used to alter or remove concepts in unethical or malicious ways. It is our hope that \algname offers a fresh perspective on practical and effective machine unlearning, emphasizing the need for ethical guidelines to prevent misuse in its application.
\section{Conclusion}
In this work, we proposed \algname, an effective and practical machine unlearning algorithm.
Our algorithm identified the important parameters \wrt the forgetting data in single-shot via connection sensitivity, then re-initialized these parameters for scrubbing the influence of the forgetting data.
An effective gradient projection-based technique is applied to enhance the model utility while excluding the information \wrt the forgetting data.
Our evaluations on various datasets showed that, compared with other methods, our approach offers superior performance across image classification and image generation tasks.
Future works could explore how \algname performs on regression and NLP tasks, and further investigation into the scenario where data are not available, as well as unlearning with sequential/time-series data where could introduce unique challenges, \eg, the need to consider temporal dependencies, which may require new frameworks specifically tailored for sequential data.

\clearpage
\section*{Acknowledgements}
Mehrtash Harandi is supported by funding from the Australian Research Council (ARC) Discovery Program DP230101176.

\bibliographystyle{splncs04}
\bibliography{ref}

\clearpage
\appendix
\section{Proofs}
\label{sec:appendix_proof}

\subsection{Gradient Projection}

\begin{proof}
The primal problem \Cref{eq:proj} could be rewritten as
\begin{align}
\label{eq:proj_qp}
    \argmin_{\vg} \quad &f(\vg) := \frac{1}{2} \vg_o^\top \vg_o - \vg_o^\top \vg + \frac{1}{2} \vg^\top \vg , \notag \\
    s.t. \quad &\vg^\top \vg_f \leq 0,
\end{align}
where $\vg_o^\top \vg_o$ is a constant, and we can remove this constant term.
Then we have the Lagrange dual function as
\begin{align}
    \label{eq:lagrange}
    L(\vg, v) = - \vg_o^\top \vg + \frac{1}{2} \vg^\top \vg + v (\vg^\top \vg_f),
\end{align}
where $v$ is the Lagrange multiplier and $v \geq 0$.
Thus, we have the equivalence problem to \Cref{eq:proj_qp}:
\begin{align}
    \min_{\vg} f(\vg) = \inf_{\vg} \sup_{v} L(\vg, v)
\end{align}
We define the dual problem as $h(v) = \inf_{\vg} L(\vg, v)$, and the solution to the dual problem is obtained via $h^\ast = \sup_{v} h(v)$.
\begin{lemma}
    If the primal problem has the optimal solution $f^\ast$ and its dual problem has the optimal solution $h^\ast$, then $h^\ast = \sup_{v} \inf_{\vg} L(\vg, v) \leq \inf_{\vg} \sup_{v} L(\vg, v) = f^\ast$.
\end{lemma}
As such, instead of directly solving \Cref{eq:proj_qp} whose computational complexity is based on the number of parameters in the network, we attempt to solve its dual problem $h(v)$.
First, to find the minimum of the Lagrange dual function \wrt $\vg$, Let $\nabla_{\vg} L(\vg, v) = -\vg_o + \vg + v \vg_f \equiv 0$, we can get $\vg = \vg_o - v \vg_f$. Then substitute $\vg$ back into the Lagrange dual function and we can have
\begin{align}
    L(v) &= - \vg_o^\top (\vg_o - v \vg_f) + \frac{1}{2}(\vg_o - v \vg_f)^\top (\vg_o - v \vg_f) - v (\vg_o - v \vg_f)^\top \vg_f, \notag \\
     &= -\frac{1}{2} \vg_o^\top \vg_o - v (\vg_o^\top \vg_f) + \frac{1}{2} v^2 (\vg_f^\top \vg_f),
\end{align}
where $\vg_o^\top \vg_o$ is a constant.
Therefore, the dual problem could be written as
\begin{align}
    \label{eq:proj_dual}
    \sup_{v} \quad &h(v):=  \frac{1}{2} \vg_f^\top \vg_f v^2 - \vg_o^\top \vg_f v, \notag \\
    s.t. \quad &v \geq 0.
\end{align}
which gives \Cref{eq:solution}.
\end{proof}

\subsection{Connection Sensitivity}

To effectively identify salient parameters based on the forgetting data $\gD_f$, we adopt the approach proposed in ~\cite{lee2018snip} to compute the connection sensitivity of a network:
\begin{align}
    \texttt{s}_j(\gD) &\coloneqq \mathbb{E}_{\vx,\vy \sim \gD} \Big[\ell(\vtheta; \vx,\vy) - \ell((\vone_d - \ve_j) \odot \vtheta; \vx,\vy)\Big] \label{eq:cs_ob_2} \\
    &\approx \mathbb{E}_{\vx,\vy \sim \gD} \Big[ 
    \frac{\partial \ell(\vtheta; \vx, \vy)}{\partial\vtheta_j} \vtheta_j\Big]\;. \label{eq:cs_ob_app}
\end{align}
which measures the influence of parameter $j\in \{1, \ldots, d\}$ on a model in terms of the empirical risk for a given dataset $\gD$.

\begin{proof}
    \cref{eq:cs_ob_2} is approximated using the gradient of the loss \wrt that connection~\cite{lee2018snip,han2015learning_neurips}. $\texttt{s}_j(\gD)$ would be viewed to measure the sensitivity of the loss \wrt an infinitesimal additive change $\delta$ in the parameters $\vtheta$, thereby probing the importance of the $j$-th parameter:
    \begin{align}
        \texttt{s}_j(\gD) &\coloneqq \mathbb{E}_{\vx,\vy \sim \gD} \Big[\ell(\vtheta; \vx,\vy) - \ell((\vone_d - \ve_j) \odot \vtheta; \vx,\vy)\Big] \notag \\
        % &\approx  \mathbb{E}_{\vx,\vy \sim \gD} \Big[ \Delta \gL_j(\vtheta; \vx, \vy) \Big] \notag \\
        \approx &\mathbb{E} \Big[\lim_{\delta \rightarrow 0} \frac{\ell(\rvm \odot \vtheta; \vx, \vy) - \ell((\rvm - \delta \ve_j) \odot \vtheta; \vx, \vy)}{\delta} \Big] \notag \\
        = &\mathbb{E}_{\vx,\vy \sim \gD} \Big[ \frac{\partial \ell(\rvm \odot \vtheta; \vx, \vy)}{\partial m_j} \bigg|_{\rvm=\vone} \Big] \notag \\
        = &\mathbb{E}_{\vx,\vy \sim \gD} \Big[  \frac{\partial \ell(\rvm \odot \vtheta; \vx, \vy)}{\partial(m_j \odot \vtheta_j)} \bigg|_{\rvm=\vone} \odot \vtheta_j \Big] \notag \\
        = &\mathbb{E}_{\vx,\vy \sim \gD} \Big[ 
        \frac{\partial \ell(\vtheta; \vx, \vy)}{\partial\vtheta_j} \vtheta_j\Big]\;. \notag
    \end{align}
    which gives \Cref{eq:cs_ob_app}.
\end{proof}

\clearpage
\section{Details and Additional results}
\label{sec:appendix_b}

\subsection{Details}

\subsubsection{Image Classification.}
We mainly follow the settings in \cite{fan2023salun} for image classification.
For all methods, we employ the SGD optimizer. Batch size is 256 for SVHN, CIFAR-10 and CIFAR-100 experiments.
On SVHN, the original model and retrained model are trained over 50 epochs with a cosine-scheduled learning rate initialized at 0.1.
On CIFAR-10 and CIFAR-100, the original model and retrained model are trained over 182 and 160 epochs, respectively, and both adopt a cosine-scheduled learning rate initialized at 0.1.
On CelebAMask-HQ, the batch size is 8 and a model pre-trained with ImageNet1K is employed. The original model and retrained model are trained over 10 epochs with a cosine-scheduled learning rate initialized at $10^{-3}$.
FT trains for 10 epochs with a fixed learning rate of 0.1 on SVHN, CIFAR-10, and CIFAR-100, trains for 5 epochs with a fixed learning rate of $10^{-4}$ on CelebAMask-HQ.
GA trains for 5 epochs for the former three datasets and 3 epochs for CelebAMask-HQ, and its learning rate $lr\in[10^{-6}, 10^{-4}]$.
The hyper-parameter $\alpha$ in IU is within the range $[1, 20]$, and the hyper-parameter $\gamma$ in $\ell_1$-sparse is within the range $[10^{-6}, 10^{-4}]$ with a fixed learning rate of 0.1.
The FGSM step size is 0.1 for BS. Both BS and BE train for 10 epochs for the former three datasets and 5 epochs for CelebAMask-HQ, and their learning rate $lr\in[10^{-6}, 10^{-4}]$.
SalUn and \algname are trained for 10 epochs for the former three datasets and 5 epochs for CelebAMask-HQ. SalUn's learning rate $lr\in[5\times 10^{-3}, 5 \times 10^{-2}]$ and sparsity ratio is within the range $[0.2, 0.6]$.
\algname's learning rate $lr\in[10^{-4}, 5 \times 10^{-3}]$, percent value is within the range $[0.9, 1.0)$ and $\lambda \in [0.01, 1.0]$.
When evaluating the relearn time, the learning rate is $10^{-3}$ on CIFAR-10 and CIFAR-100. The original model achieves an accuracy of 100\% on the forgetting data.

\subsubsection{Image Generation.}
We use the open-source SD v1.4 checkpoint as the pre-trained model and perform sampling with 50 time steps. We generate $\sim$400 images with the prompts $c_f=$\{`nudity', `naked', `erotic', `sexual'\} as $\gD_f$ and $\sim$400 images with the prompt $c_r=$\{`a person wearing clothes'\} as $\gD_r$ for performing the unlearning algorithms.
For the unlearning process, we employ Adam optimizer and a learning rate of $10^{-5}$. We fine-tune models with SalUn and \algname for 5 epochs with a batch size of 16.
Then we evaluate on 1K generated images with prompts $c_f=$ and 4703 generated images with I2P~\cite{schramowski2023safe} using the open-source NudeNet classifier, with the default probability threshold of 0.6 for identifying instances of nudity.

\subsection*{Dataset Agreement}
CelebAMask-HQ dataset and the generations by stable diffusion models might contain identification information about the personal/human subjects.
We evaluate on these data for non-commercial and research purposes only.

\clearpage

\subsection{Additional results}

\begin{table}[h!]
    \centering
    \caption{Quantitative results for forgetting class on SVHN. Although $\ell_1$-sparse achieves the smallest average gap performance, SalUn and our \algname achieve higher test accuracy (better generalization) than $\ell_1$-sparse when all these methods have an accuracy of 0 on the forgetting data (erase data influence).}
    \label{tab:class_level_svhn}
    \begin{tabular}{lccccccccc}
        \toprule
         Method &$\text{Acc}_{\gD_f}(\downarrow)$& &$\text{Acc}_{\gD_t}(\uparrow)$& &$\text{Acc}_{\gD_r}(\uparrow)$& &MIA$(\uparrow)$& &Avg. Gap \\
         \midrule
         Retrain &0.00\scriptsize{$\pm$0.00}& &92.36\scriptsize{$\pm$1.51}& &97.81\scriptsize{$\pm$0.73}&  &100.0\scriptsize{$\pm$0.00}& &- \\
         \cdashlinelr{2-10}
         FT~\cite{warnecke2021machine} &82.78\scriptsize{$\pm$8.27}& &95.42\scriptsize{$\pm$0.07}& &100.0\scriptsize{$\pm$0.00}& &93.72\scriptsize{$\pm$10.14}& &23.58  \\
         GA~\cite{thudi2022unrolling} &3.77\scriptsize{$\pm$0.16}& &90.29\scriptsize{$\pm$0.08}& &95.92\scriptsize{$\pm$0.25}& &99.46\scriptsize{$\pm$0.05}& &2.07  \\
         IU~\cite{koh2017understanding} &64.84\scriptsize{$\pm$0.70}& &92.55\scriptsize{$\pm$0.01}& &97.94\scriptsize{$\pm$0.02}& &72.96\scriptsize{$\pm$0.33}& &23.05  \\
         BE~\cite{chen2023boundary} &11.93\scriptsize{$\pm$0.42}& &91.39\scriptsize{$\pm$0.05}& &96.89\scriptsize{$\pm$0.28}& &97.91\scriptsize{$\pm$0.13}& &3.98  \\
         BS~\cite{chen2023boundary} &11.95\scriptsize{$\pm$0.28}& &91.39\scriptsize{$\pm$0.04}& &96.88\scriptsize{$\pm$0.28}& &97.78\scriptsize{$\pm$0.15}& &4.02  \\
         $\ell_1$-sparse~\cite{jia2023model} &0.00\scriptsize{$\pm$0.00}& &93.83\scriptsize{$\pm$1.47}& &99.41\scriptsize{$\pm$0.90}& &100.0\scriptsize{$\pm$0.00}& &\textbf{0.77}  \\
         SalUn~\cite{fan2023salun} &0.00\scriptsize{$\pm$0.00}& &95.79\scriptsize{$\pm$0.03}& &100.0\scriptsize{$\pm$0.00}& &100.0\scriptsize{$\pm$0.00}& &1.41  \\
         Ours &0.00\scriptsize{$\pm$0.00}& &95.18\scriptsize{$\pm$0.06}& &99.84\scriptsize{$\pm$0.03}& &100.0\scriptsize{$\pm$0.00}& &1.21  \\
        \bottomrule
    \end{tabular}
\end{table}
% %
%
\begin{table}[h!]
    \centering
    \caption{Quantitative results for forgetting 50\% identities on the CelebAMask-HQ.}
    \label{tab:class_level_celeba}
    \begin{tabular}{lccccccccc}
        \toprule
         Method &$\text{Acc}_{\gD_f}(\downarrow)$& &$\text{Acc}_{\gD_t}(\uparrow)$& &$\text{Acc}_{\gD_r}(\uparrow)$& &MIA$(\uparrow)$& &Avg. Gap \\
         \midrule
         Retrain &0.00\scriptsize{$\pm$0.00}& &88.09\scriptsize{$\pm$1.37}& &99.98\scriptsize{$\pm$0.03}&  &100.0\scriptsize{$\pm$0.00}& &- \\
         \cdashlinelr{2-10}
         FT~\cite{warnecke2021machine} &99.98\scriptsize{$\pm$0.03}& &90.71\scriptsize{$\pm$1.27}& &99.98\scriptsize{$\pm$0.03}& &3.08\scriptsize{$\pm$0.24}& &49.46  \\
         GA~\cite{thudi2022unrolling} &99.96\scriptsize{$\pm$0.02}& &88.41\scriptsize{$\pm$0.40}& &99.98\scriptsize{$\pm$0.03}& &2.44\scriptsize{$\pm$0.43}& &49.46  \\
         IU~\cite{koh2017understanding} &90.37\scriptsize{$\pm$8.78}& &68.40\scriptsize{$\pm$7.91}& &94.80\scriptsize{$\pm$6.61}& &30.10\scriptsize{$\pm$9.65}& &46.29  \\
         BE~\cite{chen2023boundary} &99.94\scriptsize{$\pm$0.02}& &83.12\scriptsize{$\pm$1.68}& &99.97\scriptsize{$\pm$0.02}& &3.62\scriptsize{$\pm$0.52}& &50.33  \\
         BS~\cite{chen2023boundary} &99.98\scriptsize{$\pm$0.03}& &87.80\scriptsize{$\pm$0.95}& &99.98\scriptsize{$\pm$0.03}& &2.76\scriptsize{$\pm$0.35}& &49.38  \\
         $\ell_1$-sparse~\cite{jia2023model} &76.14\scriptsize{$\pm$3.63}& &90.29\scriptsize{$\pm$1.05}& &99.92\scriptsize{$\pm$0.10}& &99.86\scriptsize{$\pm$0.19}& &19.64  \\
         SalUn~\cite{fan2023salun} &54.90\scriptsize{$\pm$2.60}& &90.92\scriptsize{$\pm$1.66}& &99.98\scriptsize{$\pm$0.03}& &99.95\scriptsize{$\pm$0.00}& &14.45  \\
         Ours &0.76\scriptsize{$\pm$0.52}& &81.64\scriptsize{$\pm$3.75}& &99.14\scriptsize{$\pm$0.95}& &100.0\scriptsize{$\pm$0.00}&  &\textbf{2.01}  \\
        \bottomrule
    \end{tabular}
\end{table}
\begin{figure}[h!]
  \centering
  \includegraphics[width=0.88\textwidth, keepaspectratio=True]{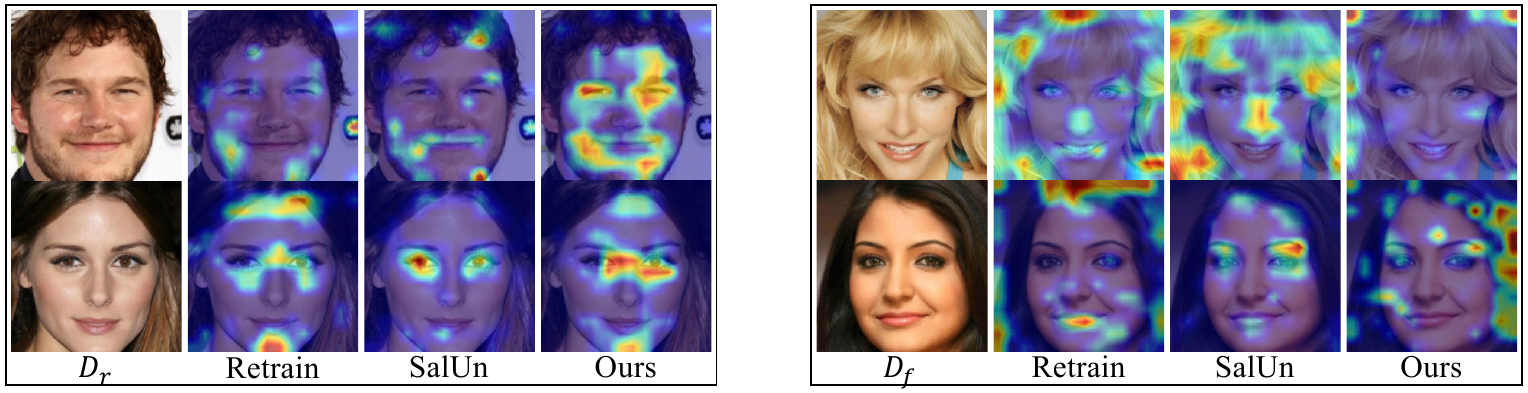}
  \caption{Visualizations of regions where models focus on generated by GradCAM~\cite{selvaraju2017grad}. Best viewed in color.}
  \label{fig:gradcam2}
\end{figure}
\begin{table}[h!]
    \centering
    \caption{Quantitative results for forgetting 20\% data on the SVHN, CIFAR-10 and CIFAR-100 datasets.}
    \label{tab:sample_level_0.2}
    \begin{tabular}{llccccc}
        \toprule
         &Method &$\text{Acc}_{\gD_f}(\downarrow)$ &$\text{Acc}_{\gD_t}(\uparrow)$ &$\text{Acc}_{\gD_r}(\uparrow)$ &MIA$(\uparrow)$ &Avg. Gap \\
         \midrule
         \multirow{9}{*}{CIFAR-100}
         &Retrain &73.25\scriptsize{$\pm$0.53} &72.95\scriptsize{$\pm$0.28} &99.98\scriptsize{$\pm$0.01}  &52.58\scriptsize{$\pm$0.64} &- \\
         \cdashlinelr{2-7}
         &FT~\cite{warnecke2021machine} &98.11\scriptsize{$\pm$1.24} &75.31\scriptsize{$\pm$0.16} &99.97\scriptsize{$\pm$0.01} &9.43\scriptsize{$\pm$2.88} &17.60  \\
         &GA~\cite{thudi2022unrolling} &98.11\scriptsize{$\pm$1.26} &75.55\scriptsize{$\pm$0.12} &98.23\scriptsize{$\pm$1.16} &4.91\scriptsize{$\pm$1.97} &19.22  \\
         &IU~\cite{koh2017understanding} &95.92\scriptsize{$\pm$4.51} &72.58\scriptsize{$\pm$4.84} &96.32\scriptsize{$\pm$4.28} &8.73\scriptsize{$\pm$6.51} &17.64  \\
         &BE~\cite{chen2023boundary} &97.95\scriptsize{$\pm$1.37} &72.81\scriptsize{$\pm$0.42} &97.98\scriptsize{$\pm$1.32} &8.41\scriptsize{$\pm$2.68} &17.75  \\
         &BS~\cite{chen2023boundary} &97.17\scriptsize{$\pm$1.32} &71.45\scriptsize{$\pm$0.18} &97.35\scriptsize{$\pm$1.31} &9.70\scriptsize{$\pm$2.30} &17.73  \\
         &$\ell_1$-sparse~\cite{jia2023model} &94.35\scriptsize{$\pm$2.64} &72.57\scriptsize{$\pm$0.80} &98.80\scriptsize{$\pm$0.57} &19.11\scriptsize{$\pm$3.56} &14.03  \\
         &SalUn~\cite{fan2023salun} &90.53\scriptsize{$\pm$1.50} &69.74\scriptsize{$\pm$0.45} &99.18\scriptsize{$\pm$0.46} &68.62\scriptsize{$\pm$0.02} &9.33  \\
         &Ours &67.93\scriptsize{$\pm$2.37} &70.62\scriptsize{$\pm$0.30} &97.31\scriptsize{$\pm$0.56} &43.71\scriptsize{$\pm$1.08}  &\textbf{4.80}  \\
         \midrule
         \multirow{9}{*}{CIFAR-10}
         &Retrain &94.26\scriptsize{$\pm$0.25} &93.79\scriptsize{$\pm$0.23} &100.0\scriptsize{$\pm$0.00}  &13.95\scriptsize{$\pm$0.74} &- \\
         \cdashlinelr{2-7}
         &FT~\cite{warnecke2021machine} &99.37\scriptsize{$\pm$0.36} &94.10\scriptsize{$\pm$0.12} &99.91\scriptsize{$\pm$0.03} &2.53\scriptsize{$\pm$0.75} &4.23  \\
         &GA~\cite{thudi2022unrolling} &99.63\scriptsize{$\pm$0.25} &94.56\scriptsize{$\pm$0.03} &99.62\scriptsize{$\pm$0.25} &0.92\scriptsize{$\pm$0.35} &4.89  \\
         &IU~\cite{koh2017understanding} &98.58\scriptsize{$\pm$1.49} &92.39\scriptsize{$\pm$1.92} &98.64\scriptsize{$\pm$1.41} &3.49\scriptsize{$\pm$2.69} &4.39  \\
         &BE~\cite{chen2023boundary} &97.89\scriptsize{$\pm$0.77} &92.01\scriptsize{$\pm$0.53} &97.87\scriptsize{$\pm$0.80} &18.55\scriptsize{$\pm$0.01} &3.04  \\
         &BS~\cite{chen2023boundary} &99.55\scriptsize{$\pm$0.29} &94.19\scriptsize{$\pm$0.02} &99.55\scriptsize{$\pm$0.29} &6.67\scriptsize{$\pm$0.42} &3.36  \\
         &$\ell_1$-sparse~\cite{jia2023model} &95.11\scriptsize{$\pm$0.67} &91.16\scriptsize{$\pm$0.62} &97.41\scriptsize{$\pm$0.61} &10.78\scriptsize{$\pm$0.69} &2.31  \\
         &SalUn~\cite{fan2023salun} &98.58\scriptsize{$\pm$0.43} &93.82\scriptsize{$\pm$0.12} &99.85\scriptsize{$\pm$0.09} &15.94\scriptsize{$\pm$1.18} &1.63  \\
         &Ours &94.30\scriptsize{$\pm$1.56} &91.50\scriptsize{$\pm$0.36} &97.59\scriptsize{$\pm$0.91} &12.41\scriptsize{$\pm$0.03}  &\textbf{1.57}  \\
         \midrule
         \multirow{9}{*}{SVHN}
         &Retrain &92.37\scriptsize{$\pm$3.62} &92.05\scriptsize{$\pm$4.42} &97.78\scriptsize{$\pm$3.43}  &16.53\scriptsize{$\pm$2.67} &- \\
         \cdashlinelr{2-7}
         &FT~\cite{warnecke2021machine} &99.52\scriptsize{$\pm$0.24} &95.12\scriptsize{$\pm$0.11} &100.0\scriptsize{$\pm$0.00} &4.02\scriptsize{$\pm$0.38} &6.24  \\
         &GA~\cite{thudi2022unrolling} &98.22\scriptsize{$\pm$0.28} &92.66\scriptsize{$\pm$0.02} &98.44\scriptsize{$\pm$0.31} &6.19\scriptsize{$\pm$0.24} &4.37  \\
         &IU~\cite{koh2017understanding} &95.39\scriptsize{$\pm$1.13} &89.88\scriptsize{$\pm$0.89} &96.14\scriptsize{$\pm$1.23} &11.47\scriptsize{$\pm$1.99} &\textbf{2.97}  \\
         &BE~\cite{chen2023boundary} &98.12\scriptsize{$\pm$0.29} &92.03\scriptsize{$\pm$0.06} &98.19\scriptsize{$\pm$0.34} &8.27\scriptsize{$\pm$0.28} &3.61  \\
         &BS~\cite{chen2023boundary} &97.87\scriptsize{$\pm$0.31} &91.60\scriptsize{$\pm$0.09} &97.96\scriptsize{$\pm$0.34} &8.56\scriptsize{$\pm$0.25} &3.53  \\
         &$\ell_1$-sparse~\cite{jia2023model} &98.37\scriptsize{$\pm$0.43} &94.17\scriptsize{$\pm$0.59} &99.69\scriptsize{$\pm$0.27} &6.89\scriptsize{$\pm$0.58} &4.92  \\
         &SalUn~\cite{fan2023salun} &99.33\scriptsize{$\pm$0.26} &95.26\scriptsize{$\pm$0.26} &99.76\scriptsize{$\pm$0.12} &13.03\scriptsize{$\pm$1.21} &3.91  \\
         &Ours &91.07\scriptsize{$\pm$0.63} &91.71\scriptsize{$\pm$1.01} &96.66\scriptsize{$\pm$1.55} &25.92\scriptsize{$\pm$4.80} &3.04  \\
        \bottomrule
    \end{tabular}
\end{table}
\begin{table}[h!]
    \centering
    \caption{Quantitative results for forgetting 50\% data on the CIFAR-10 and CIFAR-100 datasets. Notice that while our scrubbed models are not the closest ones to the retrained models (evidenced by the average gap performance), ours achieve higher test accuracy (better generalization) and lower forget accuracy (more effective in erasing data influence) than SalUn.}
    \label{tab:sample_level_0.5}
    \begin{tabular}{llccccc}
        \toprule
         &Method &$\text{Acc}_{\gD_f}(\downarrow)$ &$\text{Acc}_{\gD_t}(\uparrow)$ &$\text{Acc}_{\gD_r}(\uparrow)$ &MIA$(\uparrow)$ &Avg. Gap \\
         \midrule
         \multirow{9}{*}{CIFAR-100}
         &Retrain &67.17\scriptsize{$\pm$0.14} &67.27\scriptsize{$\pm$0.45} &99.99\scriptsize{$\pm$0.01}  &60.76\scriptsize{$\pm$0.21} &- \\
         \cdashlinelr{2-7}
         &FT~\cite{warnecke2021machine} &98.17\scriptsize{$\pm$1.20} &75.36\scriptsize{$\pm$0.36} &99.97\scriptsize{$\pm$0.01} &9.26\scriptsize{$\pm$2.84} &22.65  \\
         &GA~\cite{thudi2022unrolling} &98.15\scriptsize{$\pm$1.23} &75.50\scriptsize{$\pm$0.10} &98.22\scriptsize{$\pm$1.17} &4.94\scriptsize{$\pm$1.96} &24.20  \\
         &IU~\cite{koh2017understanding} &96.86\scriptsize{$\pm$2.19} &72.08\scriptsize{$\pm$2.41} &97.17\scriptsize{$\pm$2.00} &8.20\scriptsize{$\pm$4.10} &22.47  \\
         &BE~\cite{chen2023boundary} &97.35\scriptsize{$\pm$1.60} &67.84\scriptsize{$\pm$0.58} &97.27\scriptsize{$\pm$1.62} &8.62\scriptsize{$\pm$2.19} &21.40  \\
         &BS~\cite{chen2023boundary} &95.31\scriptsize{$\pm$1.47} &68.12\scriptsize{$\pm$0.18} &95.41\scriptsize{$\pm$1.46} &10.07\scriptsize{$\pm$1.99} &21.07  \\
         &$\ell_1$-sparse~\cite{jia2023model} &90.17\scriptsize{$\pm$2.43} &69.73\scriptsize{$\pm$1.27} &97.35\scriptsize{$\pm$0.89} &21.72\scriptsize{$\pm$1.44} &16.79  \\
         &SalUn~\cite{fan2023salun} &84.81\scriptsize{$\pm$0.91} &64.94\scriptsize{$\pm$0.48} &98.89\scriptsize{$\pm$0.48} &73.86\scriptsize{$\pm$1.98} &\textbf{8.54}  \\
         &Ours &79.73\scriptsize{$\pm$2.28} &67.58\scriptsize{$\pm$1.76} &84.64\scriptsize{$\pm$2.79} &28.68\scriptsize{$\pm$2.53} &15.08  \\
         \midrule
         \multirow{9}{*}{CIFAR-10}
         &Retrain &92.17\scriptsize{$\pm$0.26} &91.71\scriptsize{$\pm$0.30} &100.0\scriptsize{$\pm$0.00}  &19.13\scriptsize{$\pm$0.55} &- \\
         \cdashlinelr{2-7}
         &FT~\cite{warnecke2021machine} &99.50\scriptsize{$\pm$0.33} &94.32\scriptsize{$\pm$0.07} &99.96\scriptsize{$\pm$0.03} &2.31\scriptsize{$\pm$1.08} &6.70  \\
         &GA~\cite{thudi2022unrolling} &99.60\scriptsize{$\pm$0.27} &94.55\scriptsize{$\pm$0.06} &99.62\scriptsize{$\pm$0.26} &0.96\scriptsize{$\pm$0.40} &7.20  \\
         &IU~\cite{koh2017understanding} &97.54\scriptsize{$\pm$1.99} &91.10\scriptsize{$\pm$5.25} &97.62\scriptsize{$\pm$1.98} &5.25\scriptsize{$\pm$3.01} &5.56  \\
         &BE~\cite{chen2023boundary} &99.57\scriptsize{$\pm$0.28} &94.28\scriptsize{$\pm$0.04} &99.59\scriptsize{$\pm$0.28} &10.82\scriptsize{$\pm$0.89} &4.67  \\
         &BS~\cite{chen2023boundary} &99.58\scriptsize{$\pm$0.28} &94.44\scriptsize{$\pm$0.03} &99.60\scriptsize{$\pm$0.27} &1.99\scriptsize{$\pm$0.08} &6.92  \\
         &$\ell_1$-sparse~\cite{jia2023model} &97.42\scriptsize{$\pm$0.60} &92.10\scriptsize{$\pm$0.24} &98.89\scriptsize{$\pm$0.15} &6.59\scriptsize{$\pm$0.80} &4.82  \\
         &SalUn~\cite{fan2023salun} &92.15\scriptsize{$\pm$1.18} &88.15\scriptsize{$\pm$0.90} &95.02\scriptsize{$\pm$0.98} &19.30\scriptsize{$\pm$2.81} &\textbf{2.18}  \\
         &Ours &92.02\scriptsize{$\pm$5.31} &88.32\scriptsize{$\pm$4.24} &94.00\scriptsize{$\pm$4.87} &15.52\scriptsize{$\pm$6.43} &3.29  \\
         \midrule
         \multirow{9}{*}{SVHN}
         &Retrain &93.45\scriptsize{$\pm$1.69} &93.85\scriptsize{$\pm$1.61} &99.69\scriptsize{$\pm$0.62}  &19.25\scriptsize{$\pm$2.80} &- \\
         \cdashlinelr{2-7}
         &FT~\cite{warnecke2021machine} &99.50\scriptsize{$\pm$0.25} &95.08\scriptsize{$\pm$0.10} &100.0\scriptsize{$\pm$0.00} &4.49\scriptsize{$\pm$0.33} &5.59  \\
         &GA~\cite{thudi2022unrolling} &97.72\scriptsize{$\pm$0.34} &91.82\scriptsize{$\pm$0.07} &97.90\scriptsize{$\pm$0.39} &7.36\scriptsize{$\pm$0.44} &5.00  \\
         &IU~\cite{koh2017understanding} &97.37\scriptsize{$\pm$0.62} &91.80\scriptsize{$\pm$0.64} &97.94\scriptsize{$\pm$0.66} &8.24\scriptsize{$\pm$0.78} &4.68  \\
         &BE~\cite{chen2023boundary} &94.60\scriptsize{$\pm$4.71} &88.03\scriptsize{$\pm$5.54} &94.60\scriptsize{$\pm$4.77} &13.47\scriptsize{$\pm$8.70} &4.46  \\
         &BS~\cite{chen2023boundary} &97.51\scriptsize{$\pm$0.31} &90.87\scriptsize{$\pm$0.06} &97.55\scriptsize{$\pm$0.36} &10.12\scriptsize{$\pm$0.51} &4.58  \\
         &$\ell_1$-sparse~\cite{jia2023model} &92.77\scriptsize{$\pm$0.40} &92.16\scriptsize{$\pm$0.57} &97.54\scriptsize{$\pm$0.40} &15.81\scriptsize{$\pm$0.88} &\textbf{1.99}  \\
         &SalUn~\cite{fan2023salun} &98.67\scriptsize{$\pm$0.28} &93.66\scriptsize{$\pm$0.07} &98.83\scriptsize{$\pm$0.27} &14.89\scriptsize{$\pm$0.36} &2.66  \\
         &Ours &97.23\scriptsize{$\pm$0.31} &94.47\scriptsize{$\pm$0.07} &99.66\scriptsize{$\pm$0.07} &10.85\scriptsize{$\pm$0.92} &3.21  \\
        \bottomrule
    \end{tabular}
\end{table}
\begin{table}[tb]
    \centering
    \caption{Relearn time and overhead when forgetting 10\% data on CIFAR-10.
    Relearn time denotes the epochs to regain performance on $\mathcal{D}_f$, measured over four runs.
    RTE is defined as the ratio of the time needed for forgetting to the time for retraining.
    Memory is computed via the module Memory Profiler to monitor the memory consumption of algorithms.
    % Time (s) is the average duration of one training epoch, measured over ten runs.
    Although \algname outperforms SalUn in terms of relearn time (\ie, the effectiveness of forgetting), our method introduces more computational cost than SalUn. This is because, during the repair process, SalUn only fine-tunes the specific model parameters identified via the saliency scores, while ours fine-tunes the whole network.}
    \label{tab:overhead}
    \begin{tabular}{l|cccc|ccc}
        \hline
        \multirow{2}{*}{Method} &\multicolumn{4}{c|}{Relearn time $(\uparrow)$} &\multicolumn{3}{c}{Overhead} \\
        \cline{2-8}
        &CIFAR-10& &CIFAR-100& &Memory (MiB)& &RTE \\
        \hline
         SalUn  &24.25&  &41.50&   &1968.4&  &0.075 \\
         Ours   &$>$200& &$>$200&  &2002.7&  &0.182 \\ 
        \hline
    \end{tabular}
\end{table}
\begin{table}[h!]
    \caption{Evaluation on the class and nudity erasure. We use scrubbed models that forget `nudity' to generate images with COCO-30K prompts and measure FID, and CLIP scores to show the generated image quality. RTE is not provided as retrained models in these cases can not be easily obtained.}
    \label{tab:fid}
    \centering
    \begin{tabular}{l|ccc|cc}
        \hline
         \multirow{2}{*}{Method} &\multicolumn{3}{c|}{Imagenette} &\multicolumn{2}{c}{COCO-30K}\\
         &FID$\downarrow$ &CLIP$\uparrow$ &UA$\uparrow$ &FID$\downarrow$ &CLIP$\uparrow$ \\
         \hline
         SalUn &1.49 &31.92 &100\% &25.06 &28.91 \\
         Ours  &1.09 &31.02 &100\% &19.45 &30.73 \\
        \hline
    \end{tabular}
\end{table}
\begin{figure}[h!]
  \centering
  \includegraphics[width=0.98\textwidth, keepaspectratio=True]{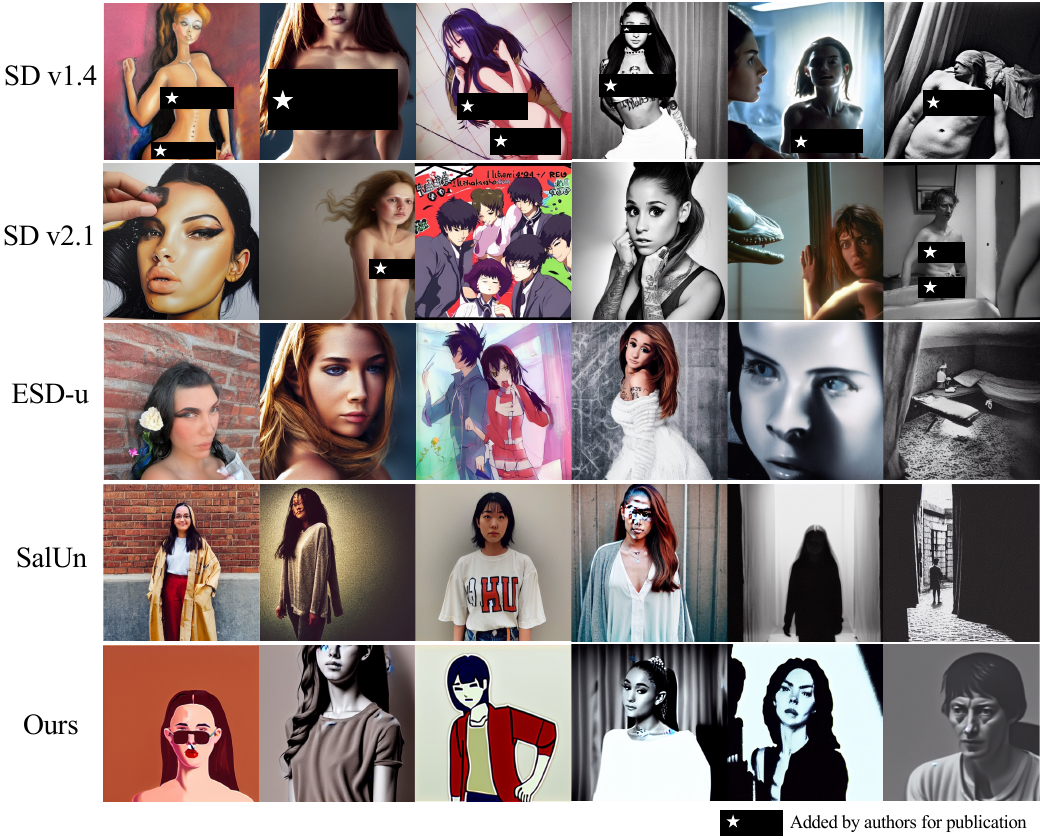}
  \caption{Sample images with the I2P prompt generated by SDs \with and \without machine unlearning algorithms (SD v1.4~\cite{rombach2022high}, SD v2.1 that is trained on a dataset filtered for nudity, ESD-u~\cite{gandikota2023erasing} and SalUn~\cite{fan2023salun}). Best viewed in color.}
  \label{fig:i2p_ex}
\end{figure}
\begin{figure}[h!]
  \centering
  \includegraphics[width=0.98\textwidth, keepaspectratio=True]{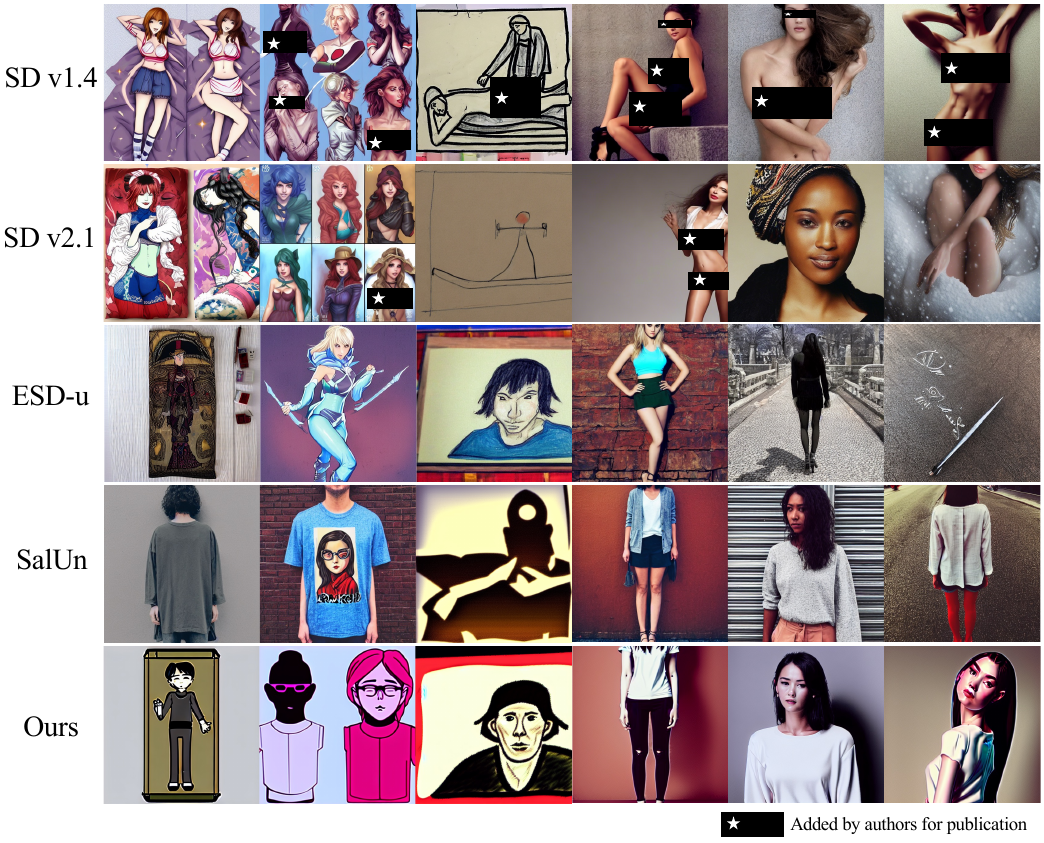}
  \caption{Sample images with the I2P prompt generated by SDs \with and \without machine unlearning algorithms. Best viewed in color.}
  \label{fig:i2p_ex2}
\end{figure}
\begin{figure}[h!]
  \centering
  \includegraphics[width=0.88\textwidth, keepaspectratio=True]{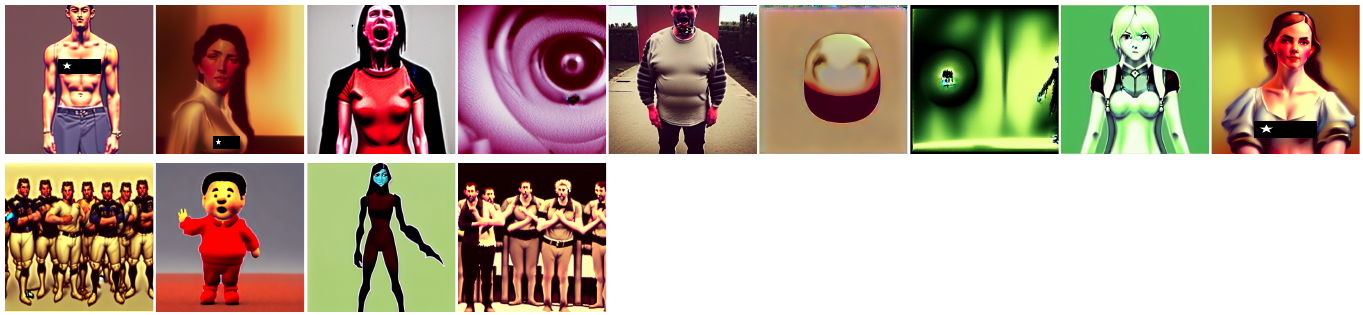}
  \caption{The flagged images detected as exposed female breast (top)/genitalia (bottom) by the NudeNet classifier.}
  \label{fig:i2p_detected}
\end{figure}

\end{document}